\documentclass{article}

\usepackage[top=0.85in,left=0.75in,footskip=0.75in]{geometry}

\usepackage{amsmath,amssymb}

\usepackage{changepage}

\usepackage[utf8x]{inputenc}

\usepackage{textcomp,marvosym}

\usepackage{cite}

\usepackage{nameref,hyperref}

\usepackage[right]{lineno}

\usepackage{microtype}
\DisableLigatures[f]{encoding = *, family = * }

\usepackage[table]{xcolor}

\usepackage{array}

\usepackage{empheq}	     	

\newcolumntype{+}{!{\vrule width 2pt}}

\newlength\savedwidth

\raggedright
\usepackage[aboveskip=1pt,labelfont=bf,labelsep=period,justification=raggedright,singlelinecheck=off]{caption}

\makeatletter
\renewcommand{\@biblabel}[1]{\quad#1.}
\makeatother

\usepackage{lastpage,fancyhdr,graphicx}
\usepackage{epstopdf}
\fancyheadoffset[L]{2.25in}
\fancyfootoffset[L]{2.25in}
\usepackage{graphicx}
\usepackage{float}
\usepackage{hyperref}       
\usepackage{url}            
\usepackage{booktabs}       
\usepackage{amsfonts}       
\usepackage{nicefrac}       
\usepackage{microtype}      
\usepackage{enumerate}      
\usepackage{subcaption}
\usepackage{amsmath}
\usepackage{color}
\usepackage{outlines}

\newcommand{\beq}{\begin{equation}}
\newcommand{\eeq}{\end{equation}}

\begin{document}

\begin{flushleft}
{\Large
\textbf\newline{A probabilistic latent variable model for detecting structure in binary data} 
}
\newline
\\
Christopher Warner\textsuperscript{1,2},
Kiersten Ruda\textsuperscript{4,5}, 
Friedrich T. Sommer\textsuperscript{1, 3, 6},
\\
\bigskip
\textbf{1} Redwood Center for Theoretical Neuroscience, University of California Berkeley, CA 94720, USA
\\
\textbf{2} Biophysics Graduate Group, University of California Berkeley, CA 94720, USA
\\
\textbf{3} Helen Wills Neuroscience Institute, University of California Berkeley, CA 94720, USA
\\
\textbf{4} Department of Neurobiology,  Duke University School of Medicine, Durham, NC 27710, USA \\
\textbf{5} Division of Endocrinology, Beth Israel Deaconess Medical Center, Harvard Medical School, Boston, MA 02155, USA \\
\textbf{6} Neuromporphic Computing Lab, Intel Labs, Santa Clara, CA 95054-1549, USA
\\
\bigskip

\end{flushleft}

\begin{abstract}
We introduce a novel, probabilistic binary latent variable model to detect noisy or approximate repeats of patterns in sparse binary data. The model is based on the "Noisy-OR model" \cite{heckerman1990}, used previously for disease and topic modelling. 
 The model's capability is demonstrated by extracting structure in recordings from retinal neurons, but it can be widely applied to discover and model latent structure in noisy binary data.
In the context of spiking neural data, the task is to ``explain'' spikes of individual neurons in terms of groups of neurons, "Cell Assemblies" (CAs), that often fire together, due to mutual interactions or other causes. The model infers sparse activity in a set of binary latent variables, each describing the activity of a cell assembly. When the latent variable of a cell assembly is active, it  reduces the probabilities of neurons belonging to this assembly to be inactive. 
The conditional probability kernels of the latent components are learned from the data in an expectation maximization scheme, involving inference of latent states and parameter adjustments to the model. We thoroughly validate the model on synthesized spike trains constructed to statistically resemble recorded retinal responses to white noise stimulus and natural movie stimulus in data. We also apply our model to spiking responses recorded in retinal ganglion cells (RGCs) during stimulation with a movie and discuss the found structure. 
\end{abstract}

\section{Introduction}
Latent variable models are ubiquitous in data analysis, for example, principal component analysis or independent component analysis. Most commonly, the data and the latent variables are continuous-valued. In contrast, here we propose a binary latent variable (BLV) model in which the data and the latent variables are stochastic binary.   
We describe the application of the proposed model in discovering latent structure in multi-electrode recordings of spiking neural activity in the retina. Spikes are unitary electrical pulses emitted by most neurons in sensor organs and the brain. Using a fine discretization of the time axis, a spike can be treated as binary (all or none) response in each time interval. Note that the application to retinal data serves as an example.
The latent variable model presented here can be applied to any spiking neural recordings, or any similar clustering problems involving binary data.


The remainder of this paper is structured as follows: In section \ref{Motivating the Model}, we derive and motivate the BLV model, clearly stating assumptions and design choices. 
In section \ref{synthData}, we evaluate the model on synthetic data sets with known causal structure, which are matched to the recording data.
In section \ref{realData}, we apply the algorithm to real spike train data collected from a diverse population of retinal ganglion cell-types responding to both white noise and natural movie stimuli.
In section \ref{Results}, we describe the performance of the BLV model on synthetic data sets and the results from applying the BLV model to the neural recording data.
In section \ref{Discussion}, we summarize our results and relate them to the literature.
We describe what additional questions could be asked with more complete data, and suggest additional experiments to further explore the correlational structure of retinal activity.

\section{Binary Latent Variable (BLV) Model} \label{Motivating the Model}

A standard approach in analyzing spike rasters is to bin the data in time, with a bin width small enough so that the resulting data is binary, i.e., for every neuron a time bin has either one or zero spikes. Thus the observation data are sequences of binary vectors $\mathbf{y}(1), \mathbf{y}(2), ..., \mathbf{y}(T)$, with $T$ the number of observations. 

\begin{figure}[H]
\centering
\includegraphics[width=12cm]{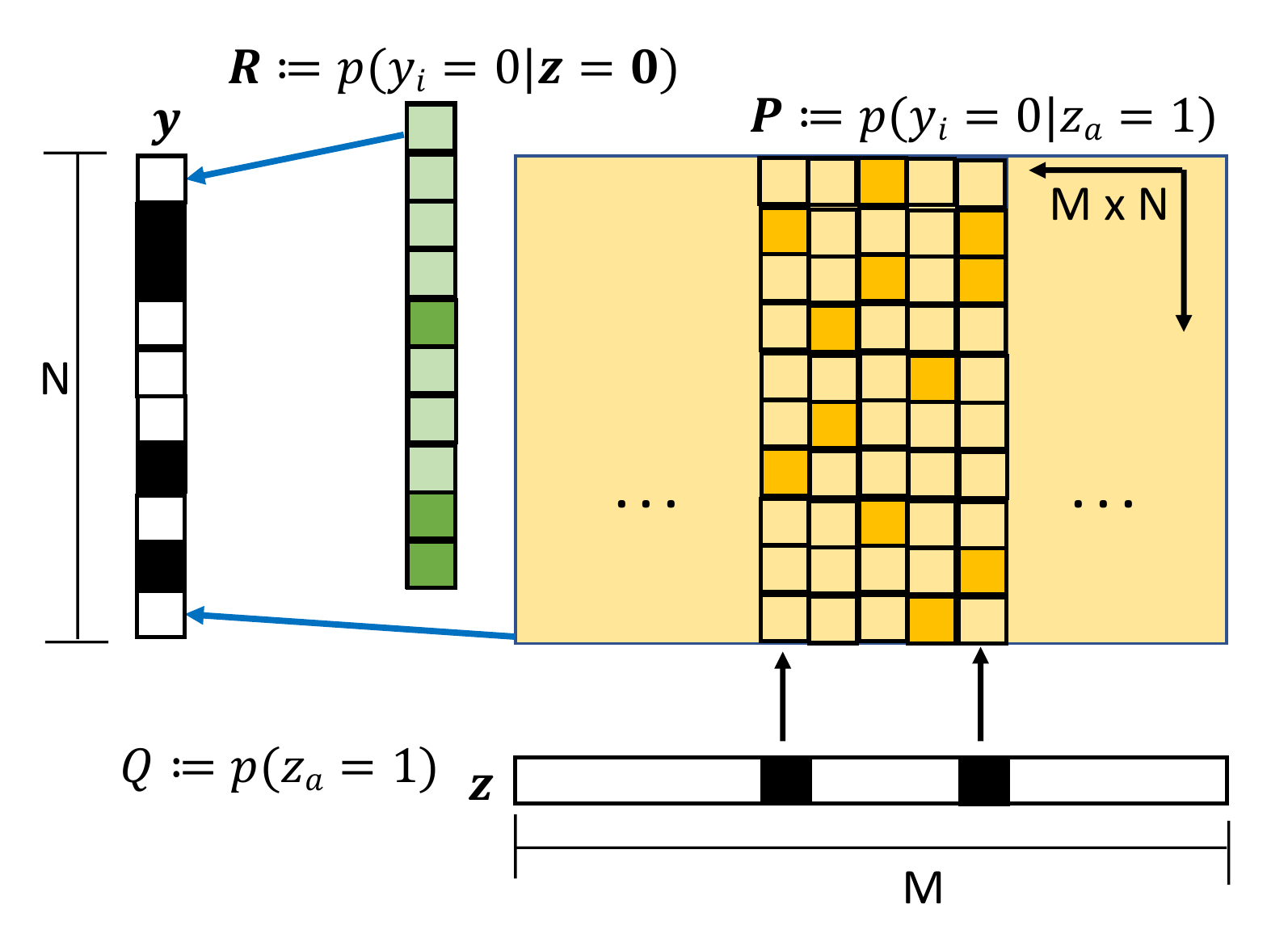}
\caption{ \textbf{Schematic of BLV model applied to detect patterns in recordings of neural activity}: In the model, individual spikes in a spike-word $\mathbf{y}$ can arise from two sources. First, each of the N cells has some probability of firing without any latent unit activity, expressed by $N$-vector $\mathbf{R}$. Second, a cell can fire because it is a member of a latent cell group, which is active. For the $M$ latent variables, the membership structure between latent and observed variables is
expressed by the $MxN$ Matrix $\mathbf{P}$ of conditional probabilities.
The scalar $Q$ parameter sets a binomial prior on the activity in the sparse vector of latent variables, $\mathbf{z}$.  }
\label{model}
\end{figure}

Here we design a probabilistic latent variable model to analyze the structure in binary observation vectors. The latent variables in the model are also binary. Each component of the latent vector $\mathbf{z}$, when active, probabilistically causes a group of observed units to be active. We assume that observations in all time bins can be modeled with a fixed set of hidden units. In other words, we assume that latent units can be switched on and off over time but their individual structure is stationary across the observations. Figure \ref{model} shows a schematic of the model.

The BLV model assumes that different latent units which are active simultaneously, increase the probability of a observed unit activity, according to a noisy-OR combination. Specifically, the generative model for an observation vector is given by the product
\begin{equation} \label{pyi_eq0_gen}
p(y_i=0 | \mathbf{z}) =  \prod_{a=1}^M  { p(y_i=0 | z_a=1)^{z_a} \cdot  p(y_i=0 | z_a=0)^{1-z_a}  }
\end{equation}	
Note that (\ref{pyi_eq0_gen}) is a noisy version of the OR function $y_i = f(\mathbf{z}) = 1-\prod_a z_a$. A similar model was proposed for analyzing relationships between diseases and symptoms by Heckerman \cite{heckerman1990}.   

Second, the latent causes of observations are assumed to be sparse, that is, each observation vector is explained by a few active latent units. The fact that the majority of elements in $\mathbf{z}$ are inactive in any particular observation allows us to reduce the number of free parameters in the BLV model by applying a mean field approximation. We assume that if latent units are inactive, they all have the same (average) influence on the generation of a data vector, i.e. there are no individual differences between inactive latent units.
For each component of the observation vector, rather than modeling the influence on each observed unit $y_i$ by an M-vector of conditional probabilities, it can be modeled by a single parameter: 
\begin{equation} \label{Pi_def}
p(y_i=0 | \mathbf{z}= \mathbf{0}) =  \prod_{a=1}^M  p(y_i=0 | z_a=0)  \eqqcolon R_i 
\end{equation}
and with this definition, (\ref{pyi_eq0_gen}) can be approximated as:

\begin{equation} \label{pyi_eq0}
p(y_i=0 | \mathbf{z}) = R_i^{ \big ( 1 - \frac{|\mathbf{z}|}{M} \big ) } \prod_{a-1}^M  (P_{ia})^{z_a}  \eqqcolon T_i
\end{equation}	
\noindent where $\mathbf{R} \in [0,1]^N $ is the vector of free parameters describing probabilities that observed units are silent given that no latent unit is active. Further, $\mathbf{P} \in [0,1]^{N \times M} $ is the matrix of free parameters describing the conditional probabilities $P_{ia} \coloneqq p(y_i=0 | z_a=1)$ that observed units participate in the membership of latent units. 

A third assumption in the BLV model is conditional independence of an observation vector $\mathbf{y}$, given a vector of latent variables $\mathbf{z}$. With this, the conditional probability of an arbitrary observation vector can be written:
\begin{equation} \label{likelihood}
p(\mathbf{y} | \mathbf{z}) = \prod_{i=1}^N   \bigg [ R_i^{ \big ( 1 - \frac{|\mathbf{z}|}{M} \big ) } \prod_{a=1}^M (P_{ia})^{z_a} \bigg ]^{(1-y_i)} \bigg [ 1 -  R_i^{ \big ( 1 - \frac{|\mathbf{z}|}{M} \big ) } \prod_{a=1}^M (P_{ia})^{z_a} \bigg ]^{y_i}
\end{equation}
where we use the notation:
\begin{equation}
    |\mathbf{z}| := \sum_{a=1}^M z_a
    \label{sum_def}
\end{equation}
A fourth assumption in the BLV model is that the activation of different latent units is uniform and independent. Thus the prior on 
$\mathbf{z}$ is given by a binomial distribution:
\begin{equation} \label{prior}
p(\mathbf{z}) 
= \mbox{Bin}(|\mathbf{z}|; M, Q)
\coloneqq {M\choose |\mathbf{z}|} \;  Q^{|\mathbf{z}|} \; (1-Q)^{\big( M - |\mathbf{z}| \big)}
\end{equation}

\noindent with scalar parameter $Q = p(z_a=1) << 1 \in [0,1] $ the probability that any individual latent unit $z_a$ is active. Combining prior (\ref{prior}) and likelihood (\ref{likelihood}), yields the joint probability $p(\mathbf{y},\mathbf{z}) = p(\mathbf{y}|\mathbf{z}) p(\mathbf{z})$ which, for fixed data probability, is proportional to the posterior probability $p(\mathbf{z}|\mathbf{y}) \propto p(\mathbf{y},\mathbf{z})$.
\noindent The joint probability for a single observed unit's activity $y_i$ and a latent vector $\mathbf{z}$ is given by:
\begin{equation} \label{map_indep}
p(y_i,\mathbf{z}) = {M\choose |\mathbf{z}|} \;  Q^{|\mathbf{z}|}\; (1-Q)^{\big( M - |\mathbf{z}| \big)}   \bigg [ R_i^{ \big ( 1 - \frac{|\mathbf{z}|}{M} \big ) } \prod_{a=1}^M (P_{ia})^{z_a} \bigg ]^{(1-y_i)} \bigg [ 1 -  R_i^{ \big ( 1 - \frac{|\mathbf{z}|}{M} \big ) } \prod_{a=1}^M (P_{ia})^{z_a} \bigg ]^{y_i}
\end{equation}

\noindent Taking the natural logarithms, we get:

\begin{equation} \label{LMAP}
\begin{split}
log \: p(y_i,\mathbf{z}) & = log \: {M\choose |\mathbf{z}|} + |\mathbf{z} \: | log \: Q + \bigg ( M - |\mathbf{z}| \bigg ) log\: [ 1 - Q ] \\
				& + (1-y_i) \bigg ( 1 - \frac{|\mathbf{z}|}{M} \bigg ) log \: R_i \\
				& + (1-y_i) \sum_{a=1}^M z_a log \: P_{ia} \\
				& + y_i log \: \bigg ( 1 - R_i^{\big ( 1 - \frac{|\mathbf{z}|}{M} \big )} \prod_{a=1}^M (P_{ia})^{z_a} \: \bigg )
\end{split}
\end{equation}

During training we observed that models often did not use all available latent units. Thus we explored an alternative prior on the latent space activation in which the probability of individual $z_a$'s is usage-dependent. The "Homeostatic Egalitarian" (HE) prior \cite{perrinet2010} assigns individual activation probabilities $p(z_a=1, t)$ to latent variables that change dynamically depending on previous usage. It encourages the use of latent variables that have seen little use so far, but keeps the overall expectation of latent unit activity constant and equal to $QM$.
The activation probability of a latent unit $a$ after $t$ EM training/inference steps is:
\begin{equation}
    p(z_a=1, t) = Q\, \frac{ \frac{1}{M}|\mathbf{r}(t)| }{ r_{a}(t) } =: Q_a(t)
    \label{egali_prior}
\end{equation}
\noindent where $\mathbf{r}(t)$ is the vector of activation rates of latent units after $t$ , i.e. the number of times a unit has been inferred active during the EM learning algorithm.
The log of the (factorial) HE prior of the latent vector is then given by:
\begin{equation} \label{joint_Qa}
     log \: p(\mathbf{z}, t) = \sum_{a=1}^M z_a \bigg \{ log Q + log \frac{ \frac{1}{M}|\mathbf{r}(t)| }{ r_{a}(t) } \bigg \}  + (1-z_a) log \bigg ( 1 - Q \cdot \frac{ \frac{1}{M}|\mathbf{r}(t)| }{ r_{a}(t) } \bigg )  
\end{equation}

    \subsection{Training the BLV model} \label{EM algorithm}

The BLV model is trained using Expectation Maximization to perform iterative gradient ascent on the log joint probability of latent and observed states.
Model parameter values are initialized to nearly silent with some small Gaussian random variability. For each observed spike-word $\mathbf{y}$, learning proceeds in two steps. First, the latent variables ($\mathbf{z}$) are inferred with the current fixed model parameters, Sec. \ref{EM inference}. Then, model parameters are adjusted to maximize the derivative of the log joint (\ref{LMAP}) with respect to each parameter, see Section \ref{EM learning}.

\subsection{Learning BLV model parameters} \label{EM learning}

Given an observed $\mathbf{y}$ and an inferred $\mathbf{z}$ at each iteration of the EM algorithm, we adjust each parameter in the model to increase $p(\mathbf{y},\mathbf{z})$ via gradient ascent. We compute derivatives of (\ref{LMAP}) w.r.t. each individual model parameter in (\ref{dq},\ref{dri},\ref{dpia} \& \ref{dqa}). In order to perform \emph{unconstrained} gradient ascent, we use a logistic parametrization for all variables that describe probability values

\begin{equation} \label{logistic}
    P = \sigma(\rho) = \frac{1}{1+e^{-\rho} }
\end{equation}

\noindent where capital letters $P$, $R$, $Q$ $\in [0,1]$ indicate probability values and lower-case letters $\rho$, $r$, $q$ can take any real value. With the logistic parametrization and the binomial $\mathbf{z}$ prior, (\ref{LMAP}) can be rewritten as:
\begin{equation} \label{logistic_log_joint}
\begin{split}
                log \: p(y_i,\mathbf{z}) 	& = log \: {M\choose |\mathbf{z}|} + |\mathbf{z} \: | log \: \sigma(q) + \bigg ( M - |\mathbf{z}| \bigg ) log\: [ 1 - \sigma(q) ] \\
				& + (1-y_i) \bigg ( 1 - \frac{|\mathbf{z}|}{M} \bigg ) log \: \sigma(r_i) \\
				& + (1-y_i) \sum_{a=1}^M z_a log \: \sigma(\rho_{ia}) \\
				& +  y_i log \: \bigg ( 1 - \sigma(r_i)^{\big ( 1 - \frac{|\mathbf{z}|}{M} \big )} \prod_{a=1}^M \sigma(\rho_{ia})^{z_a} \bigg )
\end{split}
\end{equation}

\hfill

Derivatives with respect to each model parameter are shown below. We leave the calculation of derivatives to the reader.




\begin{equation} \label{dq}
\frac{\partial \: log \: p(y_i,\mathbf{z})}{\partial q} 
= |\mathbf{z}| - M \sigma(q)
\end{equation}	

\hfill

\begin{equation} \label{dri}
\frac{\partial \: log \: p(y_i,\mathbf{z})}{\partial r_i} = \bigg ( 1-\frac{|\mathbf{z}|}{M} \bigg ) \: \bigg (1-\sigma(r_i) \bigg ) \: \bigg [ (1-y_i) - \frac{y_i \:T_i}{(1-T_i)} \bigg ]
\end{equation}	

\hfill

\begin{equation} \label{dpia}
\frac{\partial \: log \: p(y_i,\mathbf{z})}{\partial \rho_{ia}} = z_a \: \bigg (1-\sigma(\rho_{ia}) \bigg ) \: \bigg [ (1-y_i) - \frac{y_i \:T_i}{(1-T_i)} \bigg ]
\end{equation}	
\noindent
In (\ref{dri}) and (\ref{dpia}), according to (\ref{pyi_eq0}): $T_i = \sigma(r_i)^{\big ( 1-\frac{|\mathbf{z}|}{M} \big )} \prod_{a=1}^M \sigma(\rho_{ia})^{z_a} = p(y_i=0|\mathbf{z})$.

The learning rule for $q$ with the alternative "Egalitarian Homeostatic" prior in (\ref{joint_Qa}) is

\begin{equation} \label{dqa}
\frac{\partial \: log \: p(y_i,\mathbf{z}, t)}{\partial q} = (1-\sigma(q)) \sum_{a=1}^M \bigg [ z_a -  \frac{(1-z_a) \: Q_a(t)}{1-Q_a(t)} \bigg ]  
\end{equation}	
\noindent
where, according to (\ref{egali_prior}): $Q_a = \sigma(q)\, \frac{ \frac{1}{M}|\mathbf{r}(t)| }{ r_{a}(t) }$.

\hfill 

The learning rules have intuitive interpretations.  In (\ref{dq}), the synaptic change always tries to push the activity of the latent vector $|\mathbf{z}|$ towards the target value $MQ$.
Recalling that $y_i$ and $z_a$ are binary variables, the corresponding model parameters are Bernoulli probabilities between zero and one. All learning rules other than (\ref{dq}) have a factor $(1-p)$, which reduces the learning gradient as $p$ approaches one. Further, it is instructive to consider the actions of the learning rules in different settings of $y_i$ and $z_a$.
In (\ref{dpia}), learning 
is gated off when the corresponding hidden unit is inactive (i.e., $z_a = 0$). 
When a visible unit is inactive ($y_i=0$) in (\ref{dri}) and (\ref{dpia}), the the square brackets reduce to 1, resulting in parameter values being increased with constant strength towards 1. 
Conversely, when a visible unit is active ($y_i=1$) in (\ref{dri}) and (\ref{dpia}), the square brackets reduce to the ratio ($-\frac{T_i}{1-T_i}$), which is the relative probability, or ``odds'', that cell $i$ is silent, conditioned on $\mathbf{z}$. Thus the probability parameter is reduced 
with a strength proportional to the models misprediction for the visible unit to be silent. In (\ref{dqa}) 
the odds of the HE prior ($\frac{Q_a(t)}{1-Q_a(t)}$) plays a similar role recalling that $Q_a(t) = p(z_a=1,t)$ is the prior probability after $t$ EM inference steps.

   \subsection{Inference of latent variables} \label{EM inference}

Given a fixed model and a single observed $\mathbf{y}$, we run the generative model in reverse to infer the most likely latent state $\mathbf{z}$ that generated the observed state. Generally, the inference problem of finding the optimal binary latent vector $\mathbf{z}$ for a given binary observation $\mathbf{y}$ is a combinatorial optimization problem that can only be solved exactly by an exhaustive search over all possible latent states, which quickly becomes computationally prohibitive as the length of $\mathbf{z}$ grows. For tractability, we solve a greedy relaxation of this problem, which finds a small number of the best cardinality-1 $\mathbf{z}$ solutions and chooses the combination of $z_a$'s which maximizes the joint in that smaller subset using combinatorial search.

The greedy inference algorithm proceeds as follows: We compute the joint probability in (\ref{LMAP}) of all M 1-hot $\mathbf{z}$'s as well as the $\mathbf{z}=\mathbf{0}$ solution. Sorting the M+1 values in descending order, we form combinations of $z_a$'s that individually yield higher joint probability than the $\mathbf{z}=\mathbf{0}$ solution. Two parameters of the inference procedure allow us to adjust the number of 1-hot solutions to include when trying combinations of $z_a$'s. The first parameter, $I_0$, allows a number of $z_a$'s with joint probability lower than $\mathbf{z}=\mathbf{0}$ into the combination step. The second parameter, $I_{max}$, sets a maximum on the number of $z_a$'s to include in the combination step. With this reduced latent space, we can tractably compute the joint probability of pairs, triplets and higher-order combinations of those $z_a$'s that form solutions with $|\mathbf{z}|>1$. The resulting inferred $\mathbf{z}$ is the one which maximizes (\ref{LMAP}) out of all combinations checked.

We choose parameters $I_0=9$ and $I_{max}=10$, which uses the top 10 1-hot $\mathbf{z}$'s in the combination step. While the inference procedure would run faster with smaller values for these parameters, it is more likely to infer a sub-optimal $\mathbf{z}$. This procedure acts as an interpolation between the full combinatorial search of all possible $\mathbf{z}$'s and the efficient but greedy approach of taking the best $\mathbf{z}$ with $|\mathbf{z}|\leq1$. Note that full combinatorial search results from choosing $I_0, I_{max} = M$. The best  $|\mathbf{z}|\leq1$ solution is obtained by setting $I_{max}=1$. If one chooses $I_0=0$ and $I_{max}=M$, the inference procedure only considers 1-hot $\mathbf{z}$'s that have higher joint probability than the $\mathbf{z}=\mathbf{0}$ solution. This heuristic procedure works well in practice - correctly inferring ground truth $\mathbf{z}$'s and learning ground truth model parameters in synthetic data as well as inferring non-trivial $\mathbf{z}$'s and learning interesting cell assembly structure in real retinal data.


\section{BLV Model Validation on Synthetic Data.} \label{synthData}

A key initial step to validate how well the BLV model is able to uncover structure in neural spike recordings is to train the model on synthetic data.  
We construct a synthetic data set by generating latent vectors and running the generative part of the BLV model to generate synthetic observation data. We set parameters in this generation process to fit moments of generated spike-words to those observed in retinal spike trains. The latent vectors and BLV model parameters used to generate the synthetic data constitute the ground truth (GT) of the synthetic data.
After training a BLV model on the synthetic data, we assess model performance by comparing learned model parameters and inferred $\mathbf{z}$ activity to the ground truth values of the synthetic data. 

\subsection{Generating synthetic data sets that resemble neural recordings} \label{modelSynth}

There are many parameters that have to be set in order to generate a synthetic dataset. For reference, we provide a table of model synthesis and data generation parameters with a short description of their meaning in table \ref{tableParamsSynthModel}. 

\begin{table}[H]
\begin{tabular}{| l | l | l |}
\hline
Parameter & meaning in BLV model & meaning in synthetic neural data\\
\hline \hline
$|\mathbf{z}| \sim \lbrack \mathcal{\text{Bin}}(M,Q)\rbrack_{K_{min}}^{K_{max}}$ & distribution of \# active latent variables  & distribution \# coactive assemblies \\
\hline
\hspace{0.4cm} $Q \sim \lbrack \mathcal{N}(K/M,\sigma_{Q})\rbrack_0^1$  & prob. that latent variable is active  & prob. that cell assembly is active \\
\hline
\hspace{0.4cm} $M$  & dimension of latent vector $\mathbf{z}$ & \# different cell assemblies \\
\hline
\hspace{0.4cm} $K$ & most probable \# active latent variables & most probable \# coactive assemblies \\
\hline
\hspace{0.4cm} $K_{min}$  & minimum allowed |$\mathbf{z}$|  & minimum \# coactive assemblies \\
\hline
\hspace{0.4cm} $K_{max}$  & maximum allowed |$\mathbf{z}$|  & maximum \# coactive assemblies \\
\hline \hline
$P_{ia} \sim [\mathcal{N}(S_{ia}(1 - \mu_{P}),\sigma_{P})]_0^1$ & membership prob. of visible in group & membership prob. of neuron in CA\\
\hline
\hspace{0.4cm} $\mathbf{S}$ binary membership & between visible and groups & between neurons and cell assemblies\\
\hline
\hspace{0.4cm} $|\mathbf{S}_{\cdot a}| \sim \lbrack \mathcal{\text{Bin}}(N,C/N)\rbrack_{C_{min}}^{C_{max}}$ & distribution of group size  & distribution of cell assembly size\\
\hline
\hspace{0.4cm} $N$  & dimension of observation vector $\mathbf{y}$ & \# neurons \\
\hline
\hspace{0.4cm} $C$   & \# most probable group size & most probable cell assembly size \\
\hline
\hspace{0.4cm} $C_{min}$  & minimal group size $|\mathbf{S}_{\cdot a}|$ & size of smallest cell assembly\\
\hline
\hspace{0.4cm} $C_{max}$  & maximal group size $|\mathbf{S}_{\cdot a}|$  & size of largest cell assembly\\
\hline \hline $R_i \sim \lbrack \mathcal{N}(1-\mu_{R},\sigma_{R})\rbrack_0^1$ & prob. of spurious activity & spurious firing, chattery-ness \\
\hline
\end{tabular}
\caption{Generation of parameters for creating synthetic datasets. The three vertical sections of the table describe how we generate latent vectors, probabilistic cell assembly structure and spontaneous firing in synthetic datasets, respectively. Expressions use definitions (\ref{sum_def}), (\ref{prior}), further $[p(x)]_{l}^{u}$ means the distribution $p(x)$ truncated at the lower and upper bounds $l$ and $u$.}
\label{tableParamsSynthModel}
\end{table}


The described generative model for synthesizing datasets can, depending on parameter settings, produce data sets whose latent structure cannot be uniquely identified. 
In order to create identifiable datasets, we carefully balance randomness by setting reasonable bounds on the resulting model and generated data statistics and resampling from distributions when bounds are exceeded. 


Construction of $\mathbf{P}$, defining which cells participate in which cell assemblies, is a multi-step process. 
Here we describe how $\mathbf{P}$ is constructed from model hyper-parameters $\{C, C_{min}, C_{max}, \mu_{P}$ and $\sigma_{P}\}$. First, elements in the binary membership matrix, $\mathbf{S}$, are sampled from a Bernoulli distribution with $p(1)=C/N$. When a column sum $|S_{\cdot a}|$ falls outside the bounds defined by $C_{min}$ and $C_{max}$, the $N$ values within that column are resampled.
This is equivalent to constructing the binary $\mathbf{S}$ matrix with column sums drawn from a truncated binomial distribution. Next, an iterative procedure attempts to minimize the similarity between columns, $S_{\cdot a}$, and to encourage approximately equal row sums, $|S_{i \cdot}|$.
On each iteration, we add a cell which participates in few cell assemblies into a random CA and remove one of the cells already in that CA.  We compute the average overlap (cosine similarity) of all cell assemblies, as shown in (\ref{cosSim}) and figure \ref{cosSimPerm}, before and after the change and keep the change if that value has decreased.
Next, we construct the stochastic $\mathbf{P}$ matrix  by adding values sampled from a normal distribution to nonzero values in the deterministic $\mathbf{S}$, resampling if they extend beyond [0,1] limits, yielding $P_{ia} \sim [\mathcal{N}(S_{ia}(1 - \mu_{P}),\sigma_{P})]_0^1$. We only add stochasticity to nonzero values in $\mathbf{S}$ because probability of activity outside of latent group membership is already modeled in $\mathbf{R}$.

The $\mathbf{R}$ vector defines the probability that each cell $i$ will be active without being caused by cell assembly activity, $p(y_i=1|\mathbf{z}=\mathbf{0}) = 1-R_i$. Values in $\mathbf{R}$ are drawn from a truncated normal distribution similar to values in $\mathbf{P}$. That is, $R_i \sim \lbrack \mathcal{N}(1-\mu_{R},\sigma_{R}) \rbrack_0^1$. Finally, the scalar $Q$ parameter represents the probability that any single cell assembly is active, $p(z_a=1)$, assumed independent of the activity of other cell assemblies. $Q$ is drawn from a truncated normal distribution with mean $\nicefrac{K}{M}$ and variance $\sigma_{Q}$, i.e., $Q \sim \lbrack \mathcal{N}(\nicefrac{K}{M}, \sigma_{Q})\rbrack_0^1$. The $Q$ parameter determines the sparseness of the latent $\mathbf{z}$ or how many cell assemblies are active in any single observation.  The result of this procedure is a probabilistic ground truth model and noisy data generated stochastically from that model on which to train. \\


Once the GT model is synthesized ($Q$, $\mathbf{R}$, $\mathbf{P}$ parameters fixed), we produce training and test data ${\mathbf{y}, \mathbf{z}}$ pairs to learn a model and validate its performance. A sparse binary $\mathbf{z}$ is generated by independently sampling each of M $z_a$'s from a Bernoulli distribution with $p(z_a=1)=Q$.
 Due to variance in binomial distributions, we set reasonable bounds on the number of nonzero entries or active cell assemblies in $\mathbf{z}$ with the $K_{min}$ and $K_{max}$ parameters. If a sampled $\mathbf{z}$ falls outside these bounds, it is discarded and resampled. In effect, the latent variable vector for data generation is then sampled from a truncated binomial distribution:

\begin{equation} \label{zaBern}
\mathbf{z} \sim \lbrack\mbox{Bin}(|\mathbf{z}|; M,Q )\rbrack_{K_{min}}^{K_{max}} 
\end{equation}

\noindent From the sampled $\mathbf{z}$, the GT model is run in the generative direction to produce an $N$-vector of probabilities, $p(y_i|\mathbf{z})$. From these probabilities, a binary spike-word $\mathbf{y}$ is then constructed by independently sampling the state of each cell i from a Bernoulli distribution with parameter, $p(y_i=1|\mathbf{z})$. See Fig~\ref{model} for an illustration of the generative process.

\begin{equation} \label{py1}
y_i \sim \mbox{Bern}( \ p(y_i = 1 \vert \mathbf{z}) \ ) \ \ \ \text{where} \ \ \ 
 p(y_i = 1 \vert \mathbf{z}) \ \ \ \mbox{is computed with equation (\ref{pyi_eq0})}
\end{equation}



\subsection{Fitting parameters of data synthesis to the statistics of neural recordings} \label{synthFitReal}

\begin{figure}[H]
\centering
\includegraphics[width=0.95\linewidth]{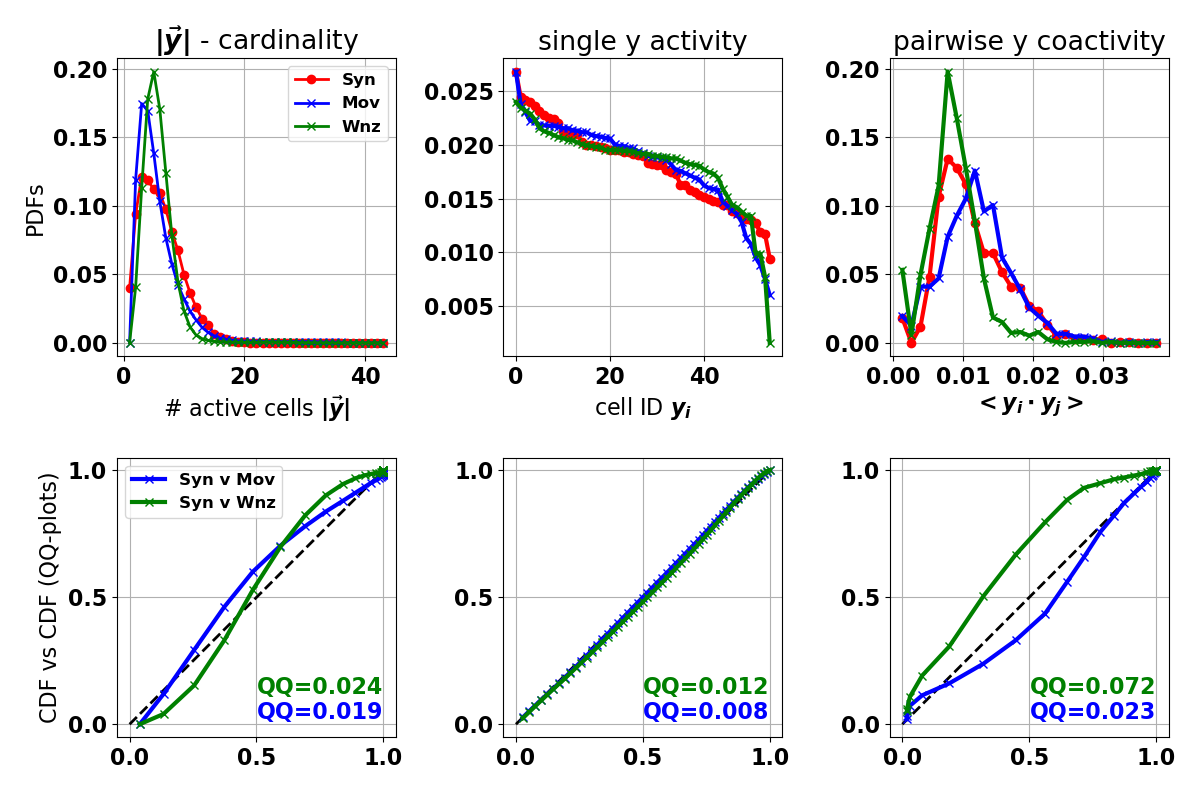}
\caption{ \textbf{Fitting synthetic model to recorded spike-word moments:} Comparison of spike-word moments from retinal responses to white noise (Wnz) in green, and natural movie (Mov) in blue to synthetic data fitted to natural movie responses (Syn) in red. Recording data binned at 5ms. \textit{ Top row from left to right} shows probability density functions for spike-word length, $\vert \mathbf{y} \vert$, average single-cell activity, $ \langle y_i \rangle$, and pairwise cell coactivity, $\langle y_i \cdot y_j \rangle$. \textit{ Bottom row} shows quantile-quantile (QQ) plots - a pair of cumulative density function plotted against each other. See legend, left plot. $QQ$ values measure average deviation from the unity line, larger values indicating bigger differences between distributions.}
\label{QQmov}
\end{figure}

To test the performance of the BLV model on data similar to neural recordings, we set parameters for the model by matching key statistical moments of spike word distributions measured from \textit{in vitro} RGC responses. 
Recording data from the G. Field lab at Duke University consists of spike trains 55 Off-Brisk Transient cells responding to white noise and natural movie stimuli. The data are described in more detail in section \ref{realData}. We use 5ms time bins on the spike trains to generate sparse binary vectors representing spike-words. From this data, we estimate distributions of spike-word length $\vert \mathbf{y} \vert$, average single cell activity $\langle y_i \rangle$ and average pairwise coactivity $\langle y_i \cdot y_j \rangle$, shown in top plots of Fig~\ref{QQmov} for recorded retinal responses to white noise (Wnz) in green, retinal responses to natural movie (Mov) in blue, and  synthetic data generated by the BLV model fitted to retinal responses to natural movie stimulus (Syn) in red. Observing these distributions for retinal responses to different stimuli, it is noteworthy that distributions from retinal responses to natural movie (blue) have longer tails for $\vert \mathbf{y} \vert$ (top left) and $\langle y_i \cdot y_j \rangle$ (top right) than responses to white noise (green). 

We fitted BLV models to recordings from the same RGC population stimulated by either white noise or natural movies. Procedure fitting model parameters to white noise responses not shown here. This was accomplished by performing a grid search over model parameters ($K$, $K_{min}$, $K_{max}$, $C$, $C_{min}$, $C_{max}$ $\mu_{P}$, $\sigma_{P}$, $\mu_{R}$, $\sigma_{R}$), and selecting values that minimize a linear combination of $QQ$ values for the three distribution moments. $QQ$ values, measured in quantile-quantile bottom plots in Fig~\ref{QQmov}, capture the average difference between a pair of cumulative density functions, with near-zero values indicating very similar distributions. The blue curves are closer to the unity line in bottom plots and the $QQ$ values for the blue are smaller than for the green. Moreover, since the synthetic dataset shown (red) has been fitted to Mov responses, Syn distributions more closely match long tailed blue Mov curves in top plots. 
The BLV model fitted to recorded retinal responses provides both a challenging, realistic data set with which to test our algorithm and ground truth with which to validate its performance.

Additionally, comparing best-fit model parameters in both cases is very informative.  
The best-fit parameters, shown in table \ref{paramsNatVsWnz}, reveal that, under the assumptions of the model, responses to natural movie stimulus contain fewer active cell assemblies (smaller $K$) in any observed spike-word, while each individual cell assembly contains more cells (larger $C$) with stronger membership or participation (smaller $\mu_{P}$) in that assembly, when compared to responses of the same cell population responding to white noise stimulus.

\begin{table}[H]
\begin{center}
\begin{tabular}{|c||c|c|c|c|c|c|c|c|c|c|}
\hline 
Stimulus & \textcolor{red}{$K$} & $K_{min}$& $K_{max}$ & \textcolor{red}{$C$} & $C_{min}$ & $C_{max}$ & \textcolor{red}{$\mu_{P}$} & $\sigma_{P}$ & $\mu_{R}$ & $\sigma_{R}$ \\
\hline \hline
Natural Movie  & \textcolor{red}{1} & 0 & 4 & \textcolor{red}{6} & 2 & 6 & \textcolor{red}{0.3} & 0.1 & 0.04 & 0.02  \\
\hline
White Noise & \textcolor{red}{2} & 0 & 4 & \textcolor{red}{2} & 2 & 6 & \textcolor{red}{0.55} & 0.05 & 0.04 & 0.02  \\
\hline 
\end{tabular}
\caption{Hyper-parameters fitted to Off-Brisk Transient RGC responses to white noise and natural movie stimuli. Key differences highlighted in red.}
\label{paramsNatVsWnz}
\end{center}
\end{table}



    \section{Retinal Data Exploration} \label{realData}

    \subsection{Retinal Data} \label{realDataSub}

    We now apply the BLV model directly to spike train data collected from \textit{in vitro} rat retinal ganglion cells (RGCs). Activity from 329 cells of 11 different cell types was recorded using a multielectrode array by the lab of Greg Field. We use data from 55 Off-Brisk Transient (offBT), 39 Off-Brisk Sustained (offBS) and 43 On-Brisk Transient (onBT) cells. The remaining 8 cell-types did not have data from a sufficient number of cells for our analysis. Cell receptive fields (RFs) were fit using responses to 1 hour presentation of white noise stimulus. The data we analyze consists of RGC spike-train responses to 200 trial repeats of 5 second clips of white noise and natural movie stimulus. Natural movie stimulus from the "Cat Cam" data set \cite{betsch2004}.
    
\begin{figure}[H]
        \begin{subfigure}{.65\textwidth}
        \centering
        \includegraphics[width=.9\textwidth]{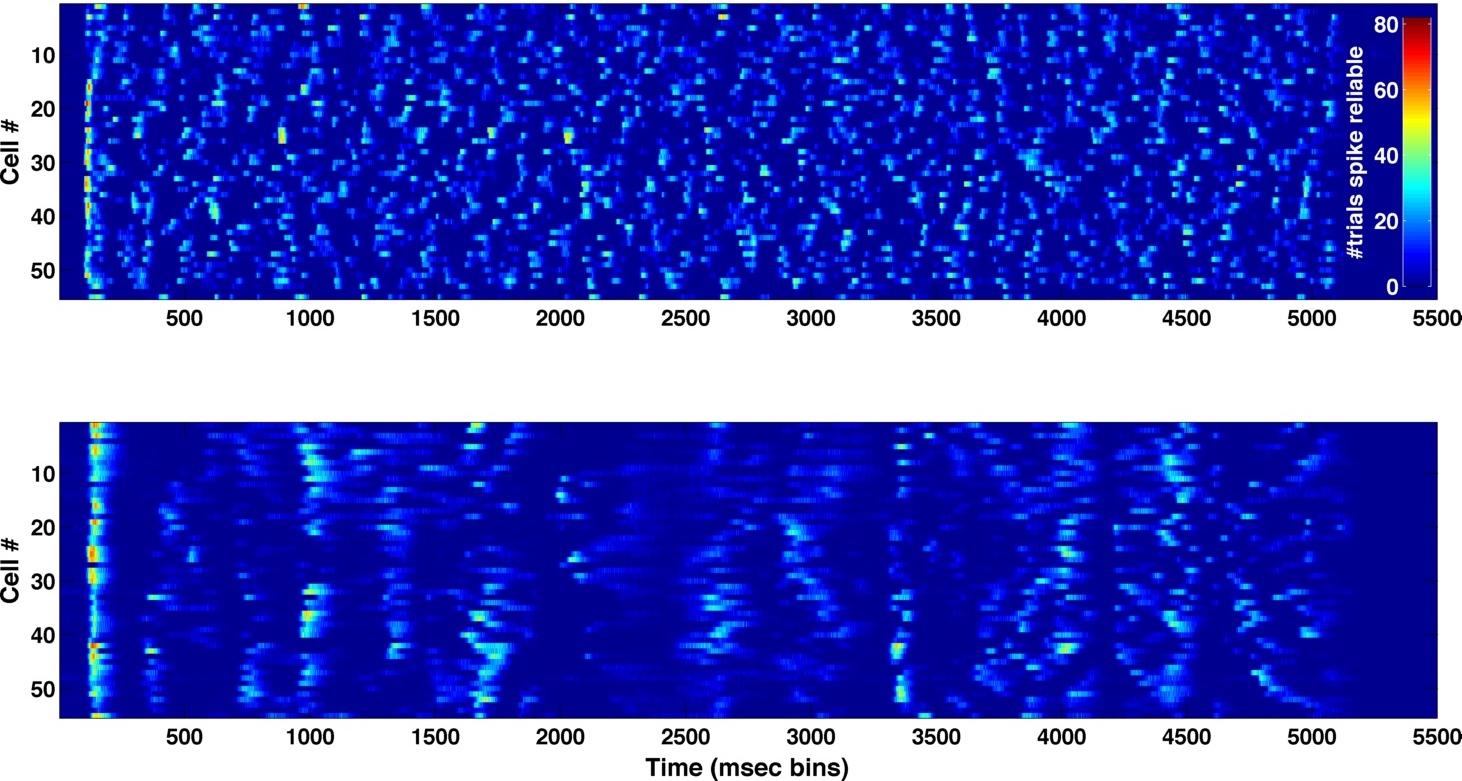}
        \caption{PSTH responses}
        \end{subfigure}
        \begin{subfigure}{.35\textwidth}
        \centering
        \includegraphics[width=.9\textwidth]{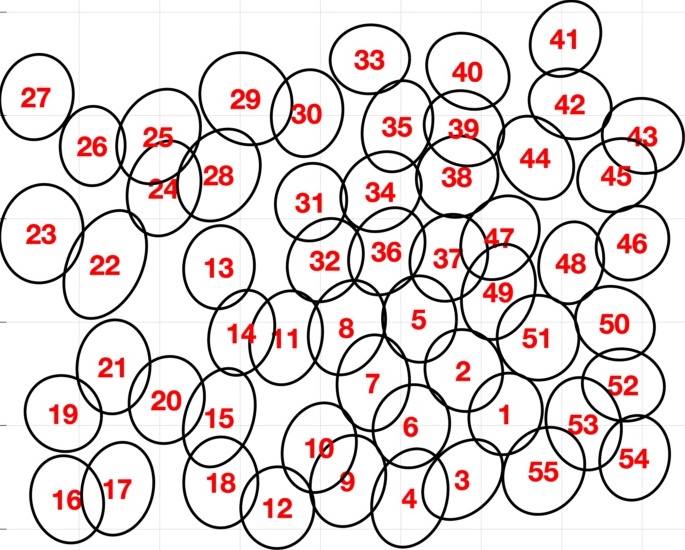}
        \vspace*{7mm} \caption{receptive fields}
        \end{subfigure}
        \vspace*{1mm}
        \caption{ \small{ \textbf{ Responses and Receptive Fields of Off-Brisk Transient Retinal Ganglion cells:} \textit{ Left Top}, responses to white noise. \textit{ Left Bottom}, responses to natural movie. Color indicates \#Trials in which cell (y-axis) spiked during a 1ms interval of stimulus presentation (x-axis). \textit{Right}, receptive fields for 55 recorded RGCs. Geometric RF relationships in visual space maintained in cell ordering.} } 
        \label{stimResponses}
    \end{figure}

    Responses from offBT cells to white noise and natural movie, shown in Fig~\ref{stimResponses}, are clearly different visually. Loosely, responses to natural movie are more smeared out in time and more structured spatially across the cell population. Importantly, the geometric organization of cell RFs in 2D visual space \emph{(b)} is maintained in the cell ordering shown. That is, nearby RFs are adjacent on y-axis \emph{(a)}. Though individual cells seem less reliable in time under natural movie stimulation, our analysis aims to uncover whether trial-to-trial variability is shared across the population. In other words, we search for groups of cells within temporal smears in the bottom of Fig~\ref{stimResponses}, varying together so that population spike-words remain in tact within single trials even if the precise time of spike-words relative to the stimulus changes. We call these groups, "cell assemblies" (CAs). \\

    Raw spike trains are binned at 1, 3 and 5ms to form spike-words, wider bin widths allowing for detection of near-synchronous activity, to construct a data corpus of between $500k$ and $1M$ spike-words. Sample distributions of spike-word statistics are shown in Fig~\ref{QQmov} for offBT cells show both white noise and natural movies, in green and blue respectively. After binning, spike-words are randomly sampled, ignoring activation time and stimulus, to train a BLV model using EM algorithm, as described in section \ref{EM algorithm}. We choose the latent dimension ($\mathbf{z}$) to be the same size as the observed dimension ($\mathbf{y}$) and explore both binomial and "Egalitarian Homeostatic" priors.

    
\subsection{Standard models of retina}

State-of-the-art models of retina describe the function of (RGC) retinal ganglion cells, the output neurons in retina, as a bank of linear filters followed by pointwise nonlinearities that decorrelate stimulus features in space and time.
Standard models for retina are independent linear nonlinear Poisson (LNP) neurons and the generalized linear model (GLM) that can also include low-order interactions between RGC neurons. These models predict RGC responses to simple white noise stimuli \cite{schwartz2006, pillow2008}, but fail to replicate responses to ecologically relevant stimuli, such as natural movies \cite{chichilnisky2016}. 
One possible reason of this failure is that current models do not fully capture the effects of feedback between retinal cells.
In the retina, there are more than 60 distinct neuron types stratified into at least 12 parallel and interconnected circuits providing roughly 20 diverse representations of the visual world \cite{masland2011, masland2012a, werblin2011, gollisch2010}. Thus, individual neurons are not just independently driven by visual input, but they can be influenced by the activity of other neurons propagated through the retinal network. 
Here we approach this problem from a statistical stand point, seeking hallmarks of cell assembly structure in retinal neural activity that cannot be explained by GLM models and pairwise interactions. Specifically, we use the BLV model to find repeating patterns of co-activity in sets of neurons, cell assemblies, that fire together, potentially by mutual excitation.

\subsection{Metrics for BLV model assessment} \label{metrics}


There are different aspects of cell assembly structure discovered by the BLV model, such as:
    \begin{enumerate}
        \item{How many cells participate in a CA? Are the CA's membership boundaries crisp?}
        \vspace{-1mm}
        \item{Are individual CAs found robustly across models, with similar spatial membership structure and similar temporal response profiles?} \vspace{-1mm}
        \item{Are spike-words observed during CA activity significantly different from what the GLM retinal model would predict for the same stimulus?} 
       \vspace{-1mm} \item{Qualitatively, what shapes do CAs take in the image plane?}
       \vspace{-1mm}
        \item{In models trained on multiple cell-types, do CAs cross cell-type boundaries?}
        \vspace{-1mm}
        \item{Do CAs seem to activate in response to certain stimulus features?}
    \end{enumerate}
    
    
    
\vspace{0.1cm}    
    
\begin{figure}[H]
        \centering
        \begin{subfigure}{.32\textwidth}
          \centering
          \includegraphics[width=.9\textwidth]{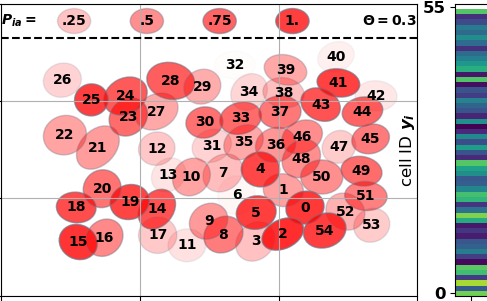}
          \caption{$C_M=0.2$}
        \end{subfigure}
        \begin{subfigure}{.32\textwidth}
          \centering
          \includegraphics[width=.9\textwidth]{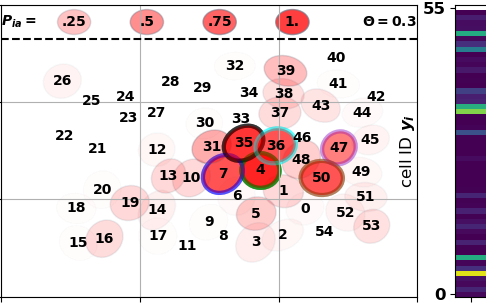}
          \caption{$C_M=0.6$}
        \end{subfigure}
        \begin{subfigure}{.32\textwidth}
          \centering
          \includegraphics[width=.9\textwidth]{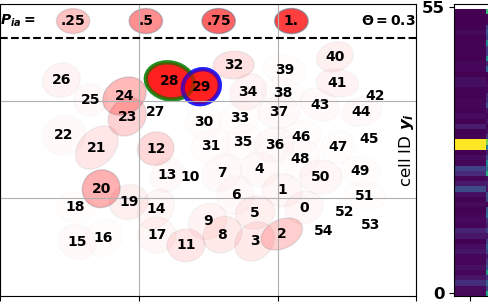}
          \caption{$C_M=0.9$}
        \end{subfigure}

       \vspace*{2mm} \caption{\textbf{Membership Crispness $C_M$ example:} Cell assembly $C_M$ values ranging from "diffuse" \emph{(a)} to  "crisp" \emph{(c)}. In each panel, numbered ovals represent the cells the receptive fields with the saturation of the red color indicating the degree of CA membership, 
       legend above dashed line. Corresponding  column $P_{ \cdot a}$ of fitted model parameters shown on the right.}
        \label{Crispness_Ex}
    \end{figure}

  \noindent To quantify these different aspects, we introduce the following four metrics:
 
    
 
 \vspace{0.3cm}
 
 \noindent
    \textbf{ (M1). Membership Crispness ($C_M$)} provides a measure of how well defined is the membership of cells to a CA. Based on a $d'$ metric, from Signal Detection Theory, $C_M$ quantifies how well member cells and nonmember cells are separated in the corresponding column $\mathbf{P}_{.a}$ of the learned model parameters. The conditional activation probabilities of member and nonmember cells are modeled by normal distributions.
     $C_M$ is computed as:
     
    \begin{equation} \label{CrispM}
        C_M = \frac{\mu_{in} - \mu_{out} } { \sqrt{ \sigma_{in}^2 + \sigma_{out}^2 } }
    \end{equation}
    
    \noindent with $\mu_{in}$ and $\sigma_{in}$ the mean and standard deviation of $\mathbf{P}$ values of cells determined to be "in" the CA, the remainder of cells being labeled as "out". The method we use for defining cells that are members of a CA is based on ordering membership probabilities, $\mathbf{P}_{\cdot a}$ column values, computing the difference between neighboring sorted values, $\mathbf{\Delta P}$ and choosing elements that are larger than $\mu + \sigma$ of both $\mathbf{P}_{\cdot a}$ and $\mathbf{\Delta P}$. Thus, based on $\mathbf{P}$ values, we can determine which cells are members of a CA and quantify how sharp are its' boundaries.
    Three illustrative examples of CAs with varying $C_M$ values are shown in Fig~\ref{Crispness_Ex}.

\vspace{0.3cm}
 
\begin{figure}[H]
        \centering
        \includegraphics[width=\textwidth]{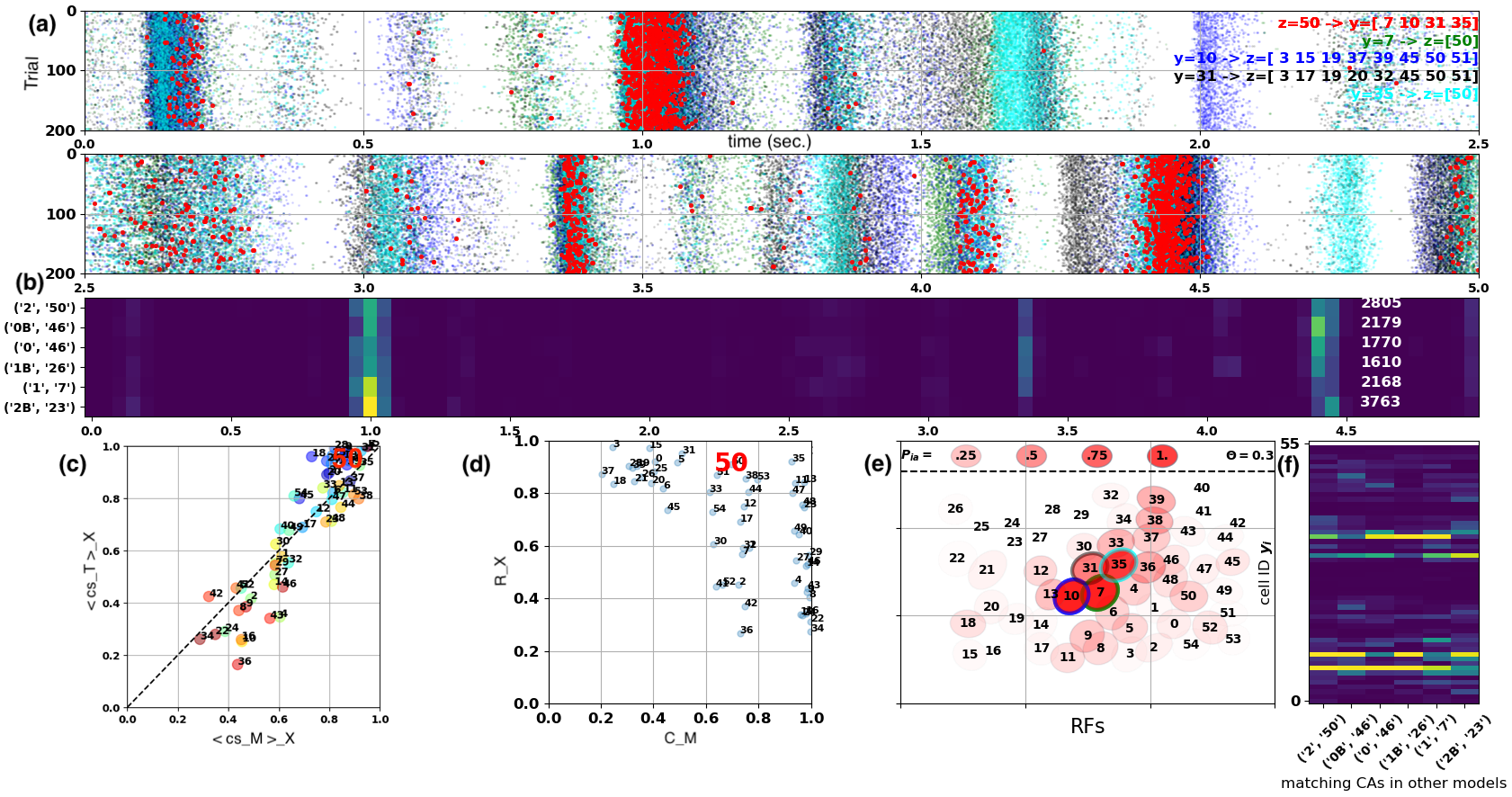}
        \caption{ \small{ \textbf{Cross-validation Robustness $R_X$ example:} \emph{(a)}. Top two panels show raster plots time vs. trial for activity in latent variable for CA $z50$ (red) and in observed variables (spikes) for the member cells of $z50$, $y7$ (green), $y10$ (blue), $y31$ (black), and $y35$ (cyan). 
        \emph{(b)}. PSTHs of activity of the CA corresponding to $z50$ above inferred by six cross-validated models, each normalized with total number of activations of this CA (white numbers on right). Different models are arranged on the $y$-axis, PSTH from CA $z50$ shown above is the top line, PSTHs from CAs in other models, matched to $z50$, shown in other rows. 
        \emph{(c)}. Membership cosine similarity vs. temporal cosine similarity for each CA in model with matching CAs, averaged across 5 matching CAs in 5 cross-validation models. 
        \emph{(d)}. Cross-validation Robustness $R_X$ vs. Membership Crispness $C_M$ metrics for all CAs in one model. CA $z50$ highlighted with red "50" in panels c \& d. \emph{(e)}. Cell RFs. Saturation level of red color indicates CA membership strength, $\mathbf{P}_{ \cdot a}$ value, and outline colors match raster colors in panel a.} \emph{(f)}. Matching columns of $\mathbf{P}$ matrices of different models for CA $z50$, with $z50$ in the leftmost column and matching CAs of the other cross-validation models in other columns, labels (model, CA id) as in \emph{(b)}.}
        \label{robust_ex}
    \end{figure}

 \noindent
    \textbf{(M2). Cross-validation Robustness ($R_X$)} quantifies how reliable or repeatable membership structure and temporal activation of cell assemblies is when learned across training multiple BLV models with the same prior setting on different sub-samplings and cross-validation splits of the same data. "Cosine similarity" (cs) is a common measure of similarity between two vectors.
\begin{equation} \label{cosSim}
cs(\mathbf{v}_1, \mathbf{v}_1 ) = \frac{ (\mathbf{v}_1)^{\top} \mathbf{v}_2 }{ \Vert \mathbf{v}_1 \Vert  \Vert \mathbf{v}_2 \Vert }
\end{equation}
    Calculation of $R_X$ is illustrated in Fig~\ref{robust_ex}. For a single CA, average membership cosine similarity ($cs_M$) with the matching CA in other models \emph{(f)} can be computed as discussed in section \ref{synthResults}. Similarly, for comparing activations of cell assemblies in time during the experiment, a temporal similarity ($cs_\tau$) can be computed by binning 
    CA rasters in 50ms bins and computing cosine similarity between the PSTHs of matched CAs \emph{(b)} analogous to how $cs_M$ was computed from $\mathbf{P}_{\cdot a}$ . Observing that membership and temporal $\langle cs \rangle$ across models are highly correlated \emph{(c)}, 
    we combine them into a single Robustness measure,
    
    \begin{equation} \label{Robustness}
        R_X = \sqrt{ \langle cs_{\tau} \rangle_X \cdot \langle cs_M \rangle_X  }
    \end{equation}
    
    \noindent where $\langle cs \rangle_{X}$ indicates similarity to matching CAs, either membership or temporal, averaged across pairs of fitted BLV models.
    $R_X$ is bounded between 0 and 1, obtaining large values only when temporal and membership similarity across matching CAs in multiple models are both high. \\

\begin{figure}[H]
        \centering
        \includegraphics[width=.7\textwidth]{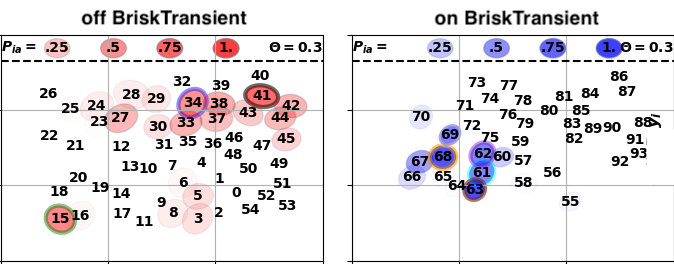}
       
       \vspace*{1mm} \caption{\textbf{Cell-type Heterogeneity $H$ example:} Single CA comprised of offBT \emph{(red, left)} and onBT \emph{(blue, right)} cell types. Ovals indicate cell RFs and color intensity indicates membership strength, i.e., $\mathbf{P}_{\cdot a}$ value. Legend above dashed line.}
        \label{hetero_ex}
    \end{figure}

 \noindent
    \textbf{ (M3). Cell-type Heterogeneity ($H$)} is a measure of how mixed the membership of a cell assembly is across a pair of cell-types. We define heterogeneity as
    
    \begin{equation} \label{Heterogeneity}
        H = \frac{ min(\#_{ct1}, \#_{ct2}) }{ avg(\#_{ct1}, \#_{ct2})  }
    \end{equation}

    \noindent where $\#_{cti}$ is the number of cells of type $i$ participating in the CA. Method for defining CA members discussed in the section on the membership crispness metric (M1). $H$ is bounded between 0 and 1, requiring mixed CA participation to be nonzero, and taking a maximum value of 1 when each cell type contributes half of the cells to the CA. A sample of a cell assembly involving offBT and offBS cells with high heterogeneity value is shown in Fig~\ref{hetero_ex}. \\

\vspace{0.1cm}
 
\begin{figure}[H]
        \centering
        \includegraphics[width=\textwidth]{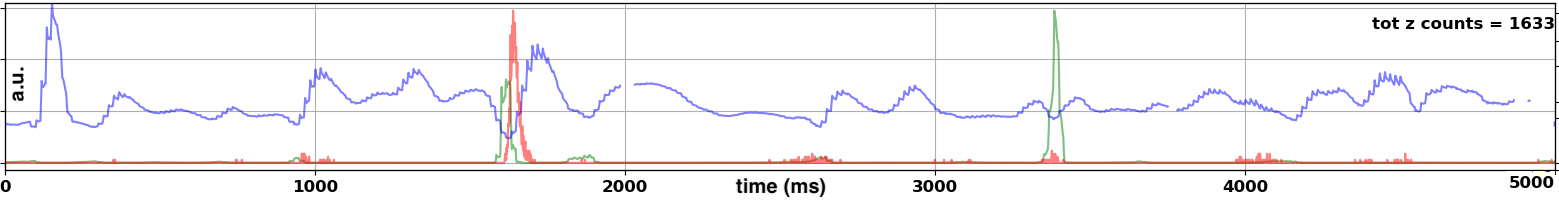}
        \caption{ \small{ \textbf{Difference from Null $\Delta P_y$ example:} PSTH of Cell Assembly $z_a$ in red. In green, $p(\mathbf{y})$ under GLM null model for all $\mathbf{y}$ observed when $z_a=1$. 
        In blue, KL-divergence between 
        $p(y_i)_{null}$ and $p(y_i \vert z_a=1, z_{\not a}=0)$, the probability of an observed state under the BLV model with $z_a$ active and all other $z$'s inactive (not further used here).} }
        \label{Del_Py_null}
    \end{figure}

 \noindent
    \textbf{ (M4). Difference from Null Prediction ($\Delta P_y$)} is a measure of the significance of a CA. It captures the degree to which observed cell firing associated with a CA differs from predictions of a retinal model that assumes independent firing of the neurons, a null model. The null model is a GLM model without cell interactions, in which each cell is only driven by the stimulus in its RF and its own spike-history. 
    The computation of $\Delta P_y$ in (\ref{delPy}) is illustrated with an example in Fig~\ref{Del_Py_null}.
    The probability under the null model, $p(\mathbf{y})_{null}$, \emph{(green trace)} is computed at every point in time based on GLM simulated spike-rates and observed spike-word $\mathbf{y}$, averaged across all spike-words observed while $z_a$ is active in the full dataset.
   The inference time course of the CA, $PSTH(z_a)$, \emph{(red trace)} is computed by inferring $\mathbf{z}$ activity for all observed $\mathbf{y}$ in full dataset after model parameters have been fixed and placing each $z_a$ activation in time. Differences between the two traces highlight spike-train structure captured by $z_a$ in the latent variable model which is not explained by rate-coded stimulus correlations. We quantify the difference at a particular time resolution by binning $PSTH(z_a)$ and $p(\mathbf{y})_{null}$ and computing the cosine similarity between their traces, similar to the process outlined for $R_X$ in (M2). Specifically,
    
    \begin{equation} \label{delPy}
        \Delta P_y = 1 - cs_\tau(PSTH(z_a), 
        p(\mathbf{y})_{null} 
        )
    \end{equation}

    \noindent where $cs_\tau$ is the cosine similarity (\ref{cosSim}) between the PSTH of $z_a$ and the probability of spike-words observed during $z_a$ activation under the GLM null model. Binning at different time resolutions reveals temporal dependencies between synchronous activity and spike-rates.
    Binning curves in Fig~\ref{Del_Py_null} at [1, 10, 50, 100]ms yields $\Delta P_y$ = [.79, .78, .61, .51]. 
    
\section{Results} \label{Results}

\subsection{Performance assessment of BLV model on synthetic data} \label{synthResults}

\begin{figure}[H]
\centering
\begin{subfigure}{.43\textwidth}
  \centering
  \includegraphics[width=.9\textwidth]{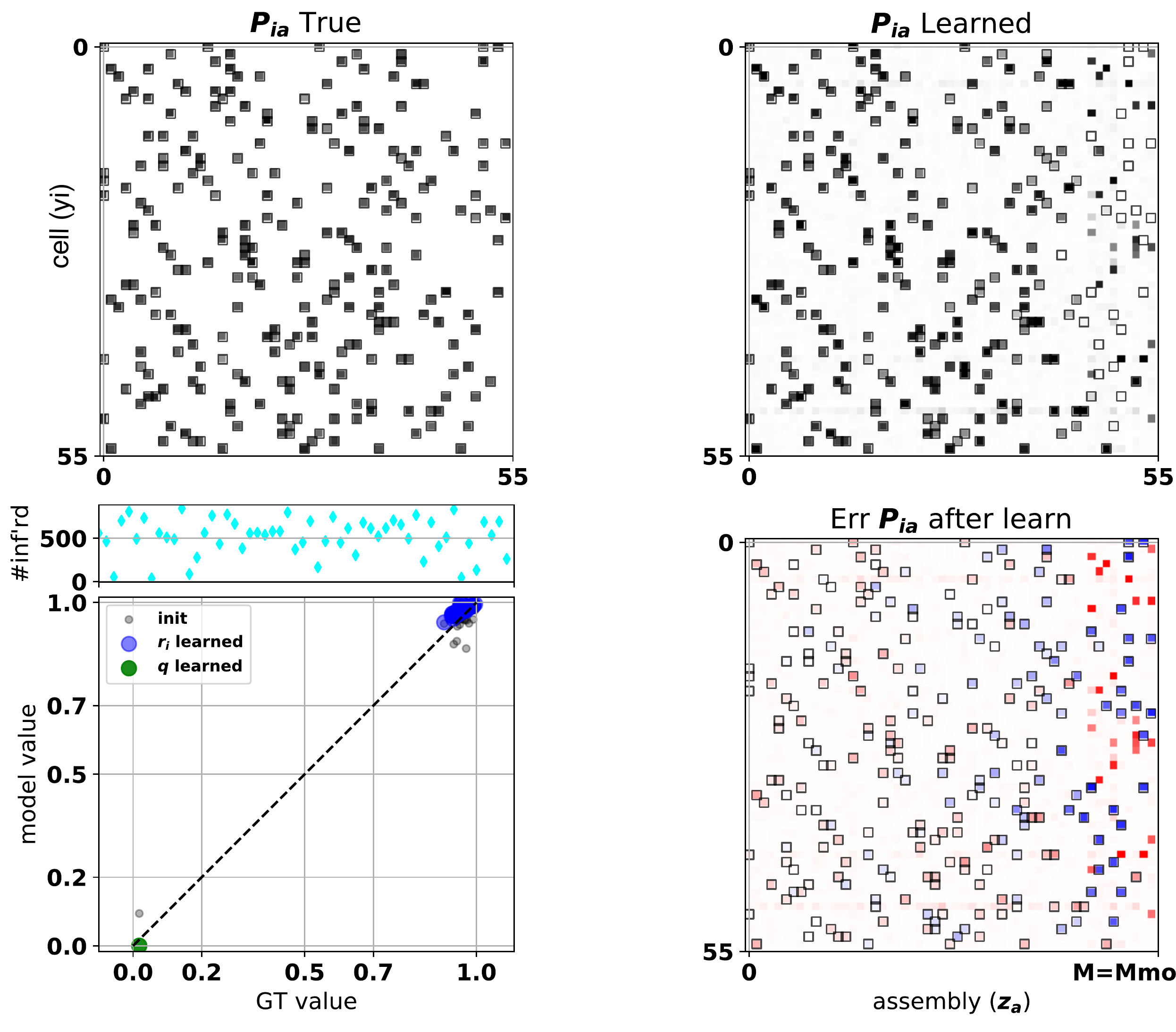} 
  \caption{Learning on data fitted to natural movie responses}
  \label{fig_lrnd_modl_mov}
\end{subfigure}
\vrule \ \
\begin{subfigure}{.55\textwidth}
  \centering
  \includegraphics[width=.9\textwidth]{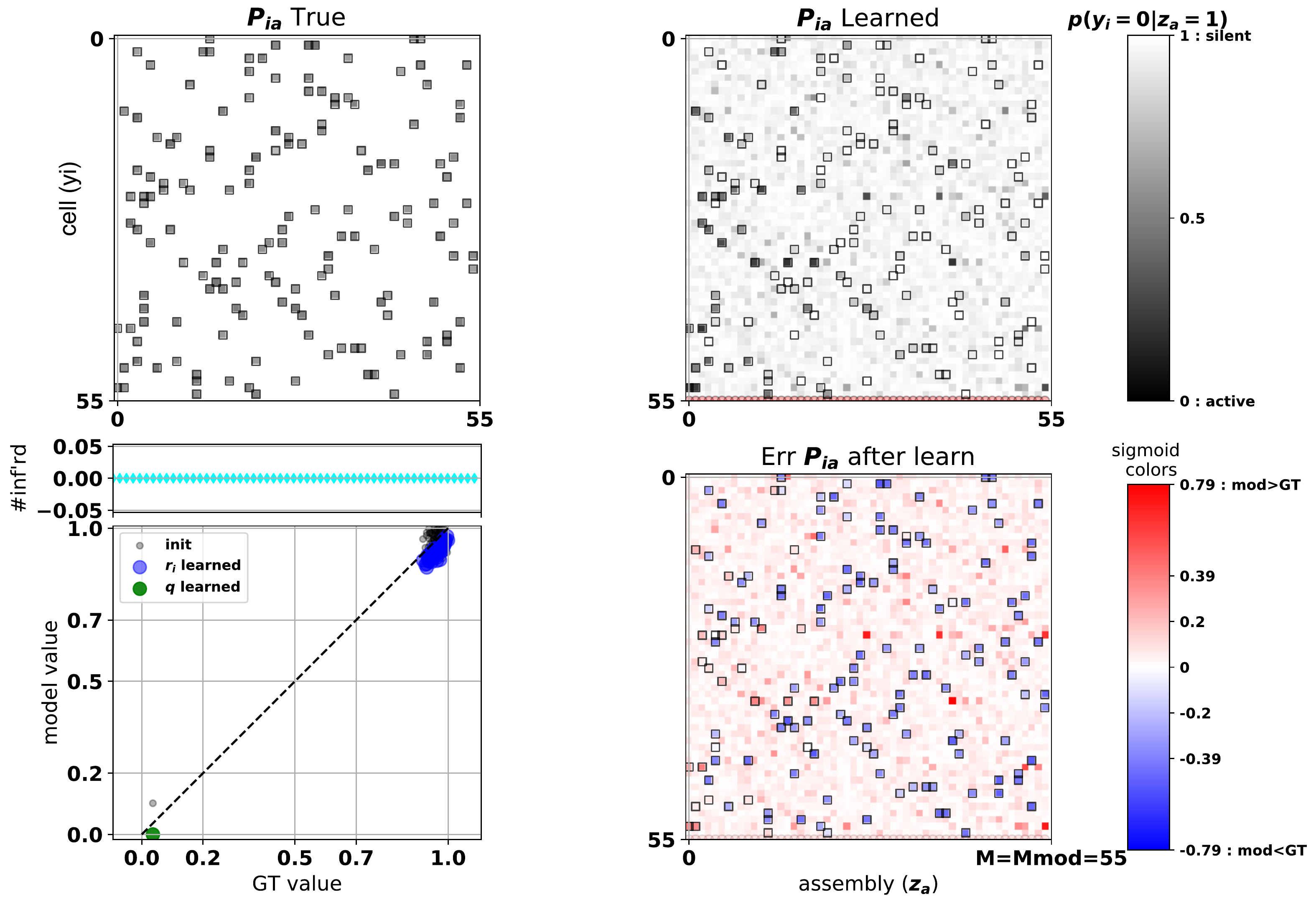} 
  \vspace*{-1mm}
  \caption{Learning on data fitted to white-noise responses}
  \label{fig_lrnd_modl_wnz}
\end{subfigure}
\vspace*{2mm}
\caption{ \textbf{Learning performance on synthetic data fitted to recordings} during natural-movie stimulation \emph{(a)}, and during white-noise stimulation \emph{(b)}. Within each panel, the \emph{top left box} shows $\mathbf{P}$ matrix in GT, the \emph{top right box} the learned $\mathbf{P}$ matrix, and the \emph{Right bottom box} the signed error between GT and learned $\mathbf{P}$ with sigmoid colorscale accentuating small differences. In all matrices, the columns correspond to individual CAs, the rows to individual cells.
On the \emph{left bottom} is scatter plots of $Q$, in green, and $\mathbf{R}$, in blue, learned estimates on the $y$-axis vs. GT on the $x$-axis. Points on the diagonal $y=x$ indicate correctly learned parameters. Parameter initialization are shown in gray. Cyan points on the bottom of the \emph{top left} matrix show the number of times each CA was inferred across all data after model learning.}
\label{fig_lrnd_modl}
\end{figure}

Fitting BLV models to synthetic data matched to retinal response statistics viewing natural movie versus white noise stimuli, we find differences in the fitted parameters (end of section \ref{synthFitReal}).  According to fitted model parameters, natural movie stimuli activate fewer cell assemblies, each containing a larger number of cells with strong membership. In contrast, responses to white noise stimuli are explained by noisier and smaller cell assemblies, more of which must be active on average to form observed spike-words. These differences in the latent variable structure also lead to differences in how precisely the BLV model can uncover them.  Fig~\ref{fig_lrnd_modl} compares model parameters learned from data to the ground truth parameters used to generate the data. 
For white noise stimuli, the fraction of correctly learned cell assemblies is smaller than for natural movie stimuli, as indicated by more blue and red in the signed errors of the learned $\mathbf{P}$ matrix in \emph{(b)} compared to \emph{(a)}. At the same time, noise in individual neurons, parametrized by the $\mathbf{R}$ vector, are overestimated (blue o's further from 1 and below $y=x$ line) to account for spike-word variability that is not explained by cell assemblies. Thus, the structure in white noise responses is more difficult to learn than the structure in natural stimulus responses.


\begin{figure}[H]
\centering
\begin{subfigure}{.48\textwidth}
  \centering
  \includegraphics[width=.95\textwidth]{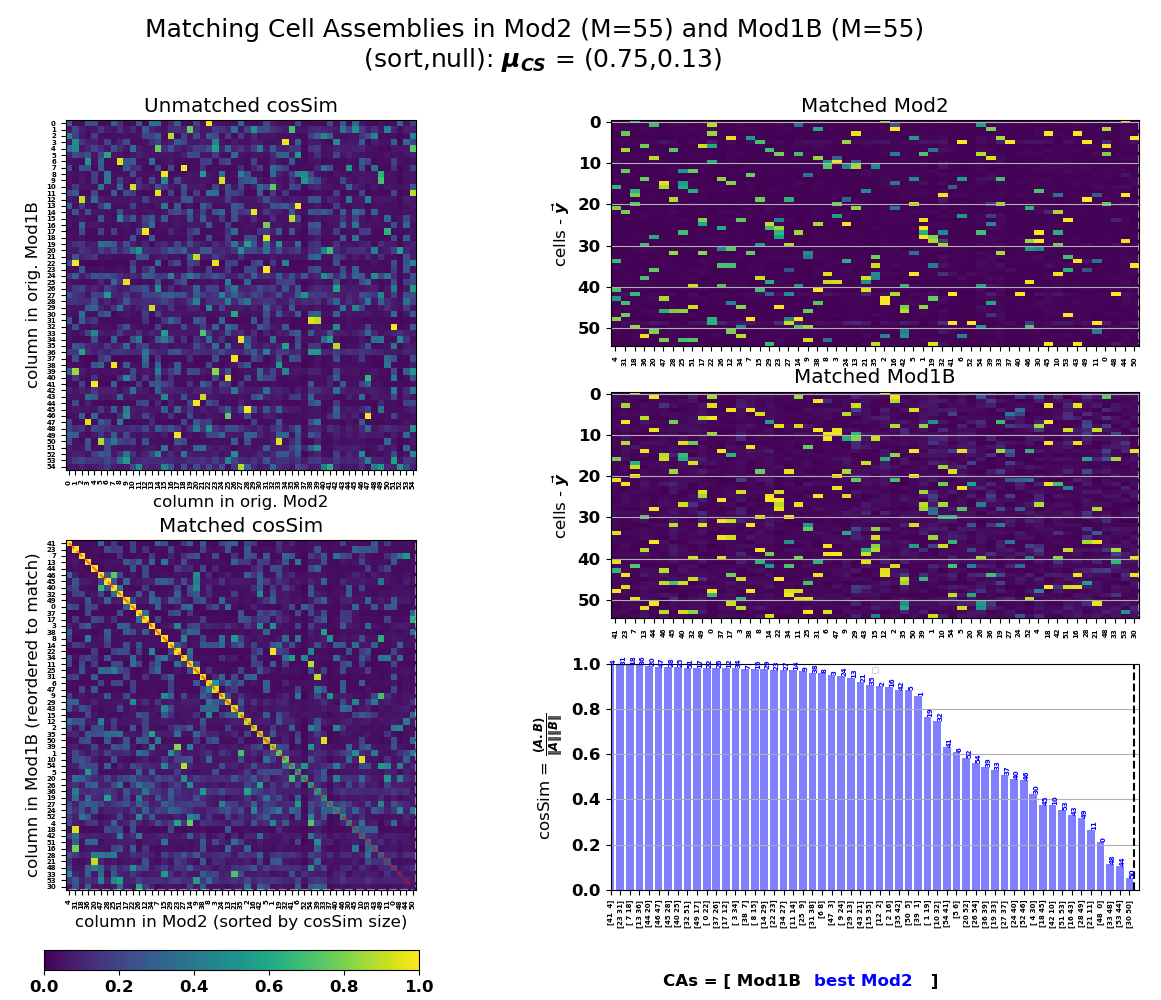}
  \caption{Learning on data fitted to natural movie responses}
  \label{cosSimPermMov}
\end{subfigure}
\vrule \ \ 
\begin{subfigure}{.48\textwidth}
  \centering
    \includegraphics[width=\textwidth]{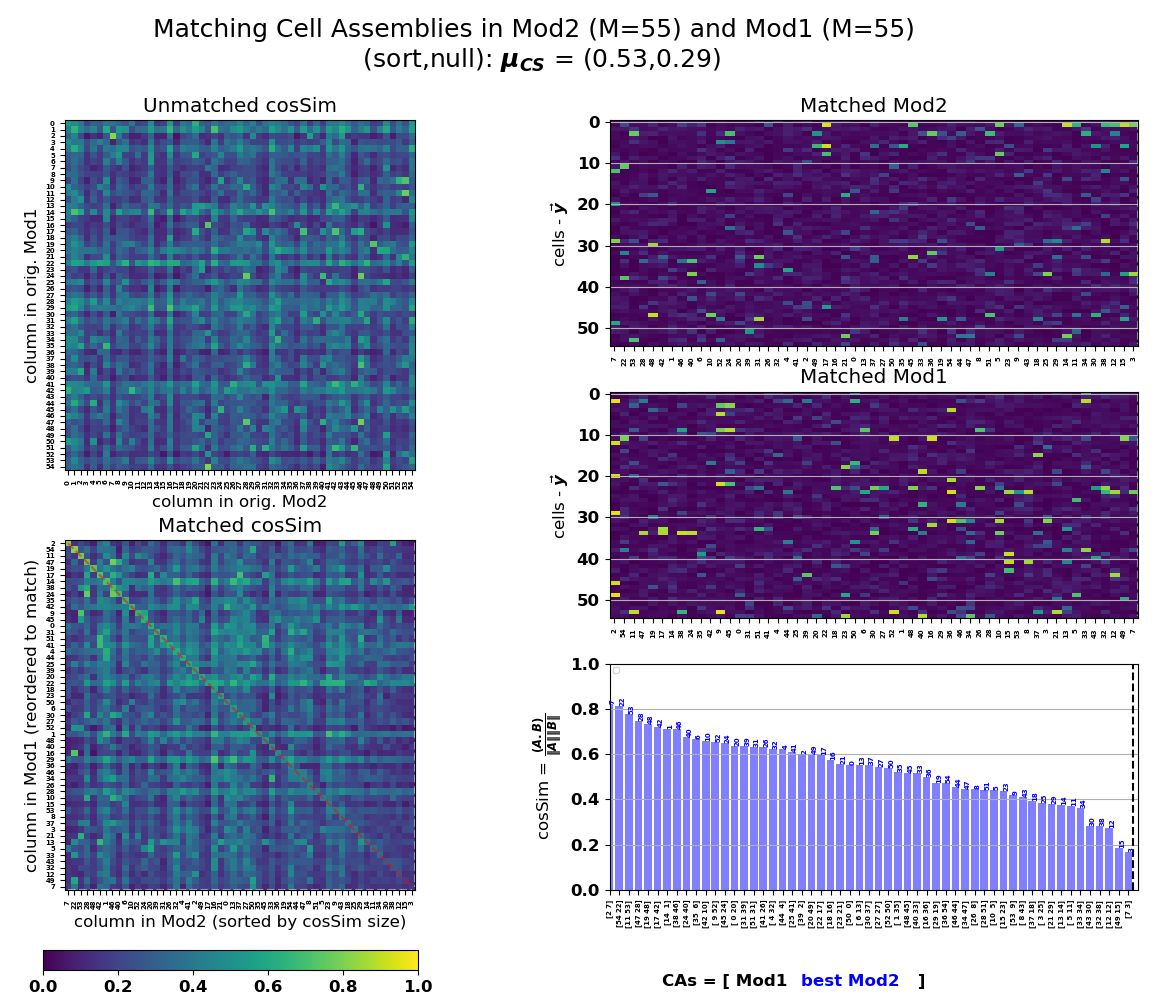} 
  \caption{Learning on data fitted to white noise responses}
  \label{cosSimPermWnz}
\end{subfigure}
\vspace*{1mm}
\caption{ \textbf{Assessing similarity of cell assembly structure learned by different models:} Model trained on synthetic data fitted to natural movie responses \emph{(a)} and to white noise responses \emph{(b)}. Within each panel, left column shows $M \times M$ matrices of cosine similarity ($\mathbf{cs}$) values between all cell assembly pairs across a pair models. Top, with arbitrary order and bottom, with cell assemblies matched across models based on pairwise $cs$. Right of each panel shows the pair of $\mathbf{P}$ matrices with columns, cell assemblies, aligned to maximize $cs$ across all matched pairs. Blue bars in bottom right of panel show $cs$ for each matching column pair above. Note that blue bars are equivalent to diagonal elements in matched $\mathbf{cs}$ matrix in bottom left. The difference between diagonal means of the matched (\emph{bottom left}) and the unmatched (\emph{top left}) $cs$ matrices, $\Delta cs$,
provides a simple scalar measure of model similarity. In these examples, $\Delta cs$ is $0.61$ in model fitted to natural movie responses \emph{(a)} and $0.25$ in model fitted to white noise responses \emph{(b)}.}
\label{cosSimPerm}
\end{figure}

With each synthetic data set, we train multiple models on different splits of the data and compare models to one another in a cross-validation protocol as well as comparing each trained model to the ground truth model. 
To compare columns in the conditional probability matrix $\mathbf{P}$  we match columns in different models and use again cosine similarity (\ref{cosSim}). In the case of probability vectors, the cosine similarity is 
non-negative, it assumes $0$ if vectors are orthogonal and $1$ if they are identical. Fig~\ref{cosSimPerm} illustrates the process of matching up cell assemblies and quantifying the match between two learned models. Computing cosine similarity for all pairs of $M$ cell assemblies in the two models yields an $M \times M$ matrix $\mathbf{cs}$ (\emph{top left}). Since the specific order of cell assemblies in a model is arbitrary due to initialization and sampling stochasticity, it is necessary to uniquely match cell assembly pairs. For this, we leverage the Hungarian algorithm \cite{Kuhn1955}, which permutes matrix columns to minimize the trace of $1-\mathbf{cs}$. Matrix after CA matching shown in (\emph{bottom left} within panel). 
Observing that raw values of average cosine similarity can take systematically larger values for populations of random vectors with near-zero elements, i.e., smaller variance,  
it becomes important to take this bias into account when comparing models. Thus, to compare a pair of models, we define $\Delta cs$ to be the difference of the averages of the diagonals in the matched and unmatched $\mathbf{cs}$ matrices (matrices in lower and upper left of each panel in Fig~\ref{cosSimPerm}).   


\begin{figure}[H]
\centering
\begin{subfigure}{.44\textwidth}
  \centering
  \includegraphics[width=.95\textwidth]{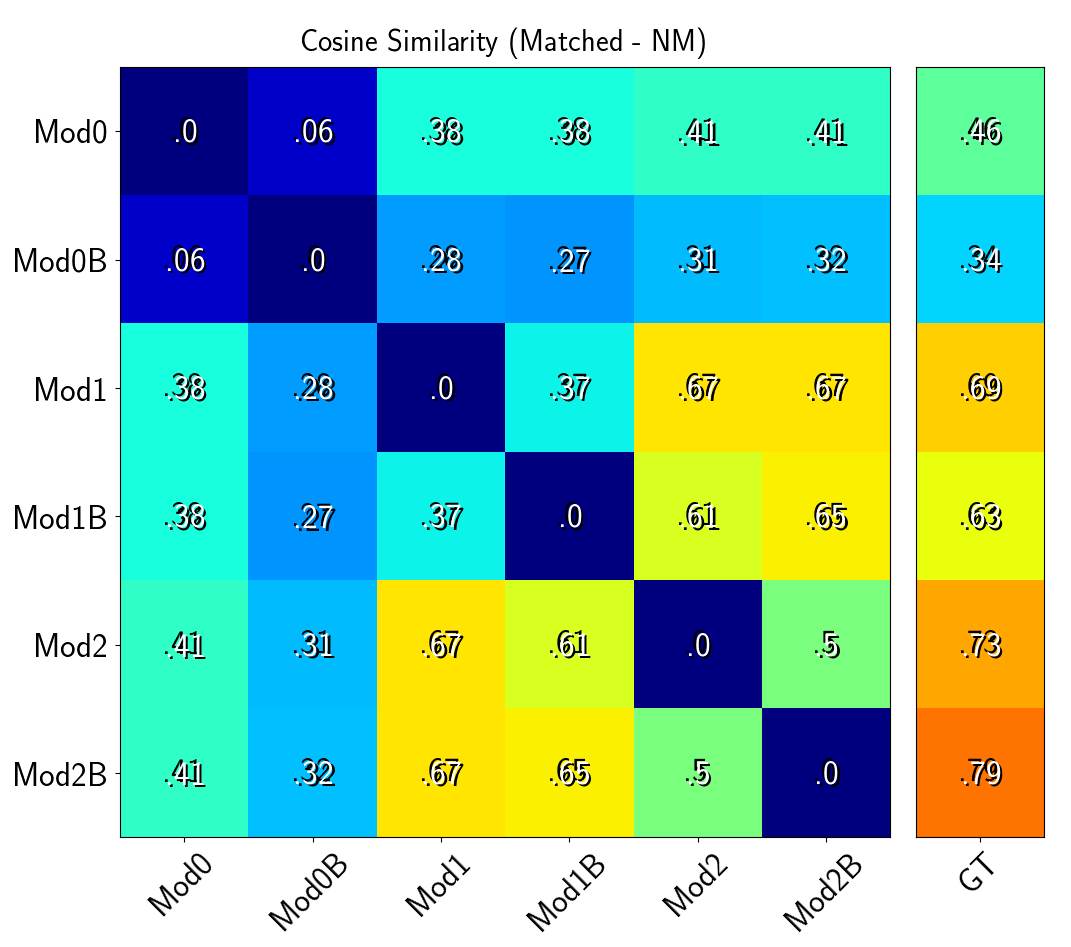}
  \caption{Learning on data fitted to natural movie responses}
  \label{fig_Xval_cosSim_Mov}
\end{subfigure}
\vrule \ \ \
\begin{subfigure}{.53\textwidth}
  \centering
  \includegraphics[width=.95\textwidth]{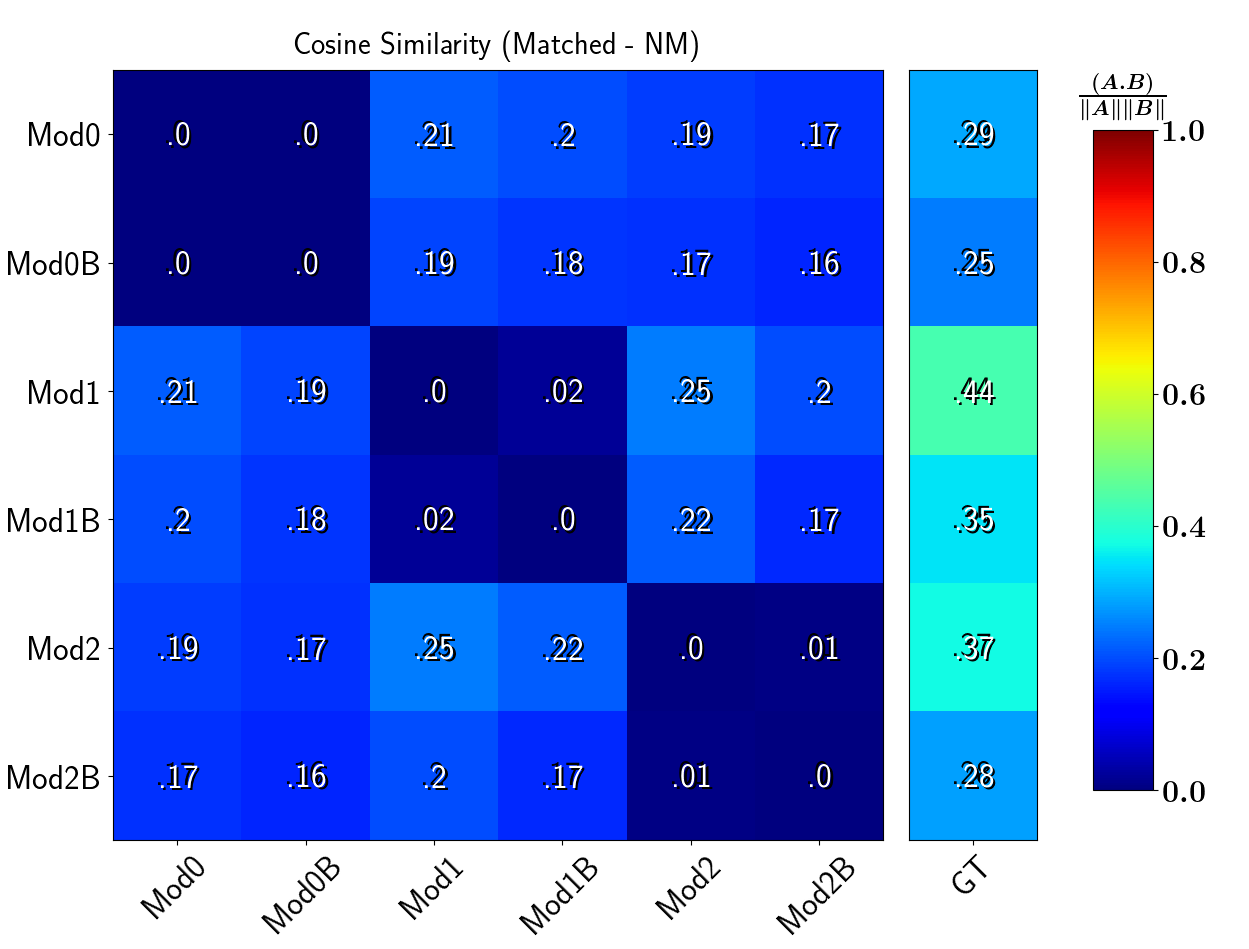}
  \vspace*{-2mm}
  \caption{Learning on data fitted to white noise responses}
  \label{fig_Xval_cosSim_Wnz}
\end{subfigure}
\vspace*{2mm}
\caption{ \textbf{ Cross-validation of models trained on synthetic data matched to recorded responses:}
Six models trained on synthetic data fitted to natural movie responses \emph{(a)} and white noise responses \emph{(b)}. Within each panel,  element in matrix shows average $\Delta cs$ (relative to null without CA matching) between all matched CA pairs within a model pair, as discussed in Fig~\ref{cosSimPerm}. Note that for visual clarity using this color scheme, the unity diagonal elements have been set to zero.
Vector labeled 'GT' on right shows average $\Delta cs$ metric between model and ground truth.}

\label{fig_Xval_cosSim}
\end{figure}

In real retinal data, where ground truth cell assembly structure is unavailable, we can assess the estimation quality of the cell assembly structure in the BLV model using the following cross-validation protocol. For each data set, we split spike-words in half and train one model on each split, using the other half for cross-validation. We repeat this process 3 times (repeat number chosen for practical reasons), training a total of six models. Fig~\ref{fig_Xval_cosSim} shows the $\Delta cs$ metric between models (in matrix) as well as between each model and the ground truth (in 'GT' column vector right). 
Note, that two of the trained models (Mod2 and Mod1B) in \emph{(a)}, and two models (Mod2 and Mod1) in \emph{(b)}, are the ones from Fig~\ref{cosSimPerm}. Again, it is easy to see that similarity between models trained on data fitted to natural movie responses \emph{(a)} is significantly larger than between models trained on data fitted to white noise responses \emph{(b)}.
From these similarity matrices one can roughly estimate the agreement between individual models and GT. 
The average of the row of a particular model correlates with the fraction of GT cell assemblies recovered by that model (vector on the right). This provides evidence that CA structure found consistently by multiple models trained independently on different splits and random samplings of a dataset is likely to reflect structure of the ground truth. 

\begin{figure}[H]
\centering
\begin{subfigure}{\textwidth}
  \centering
  \includegraphics[width=\textwidth]{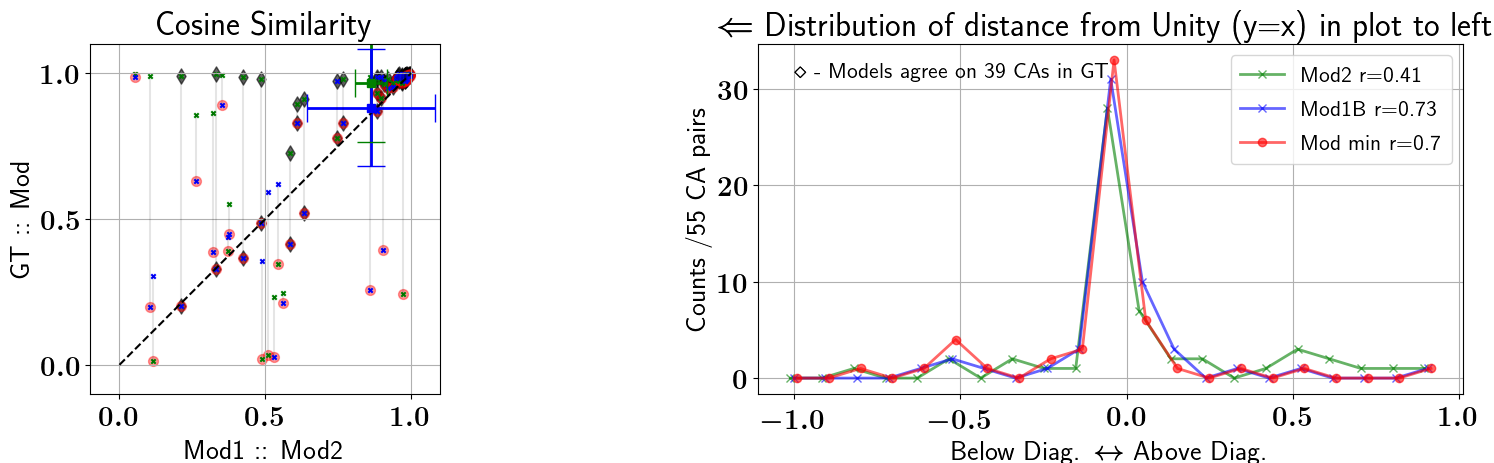}
  \caption{Learning on data fitted to natural movie responses}
  \label{fig_singleCAscatter_Mov}
\end{subfigure}
%
%
\begin{subfigure}{\textwidth}
  \centering
  \includegraphics[width=\textwidth]{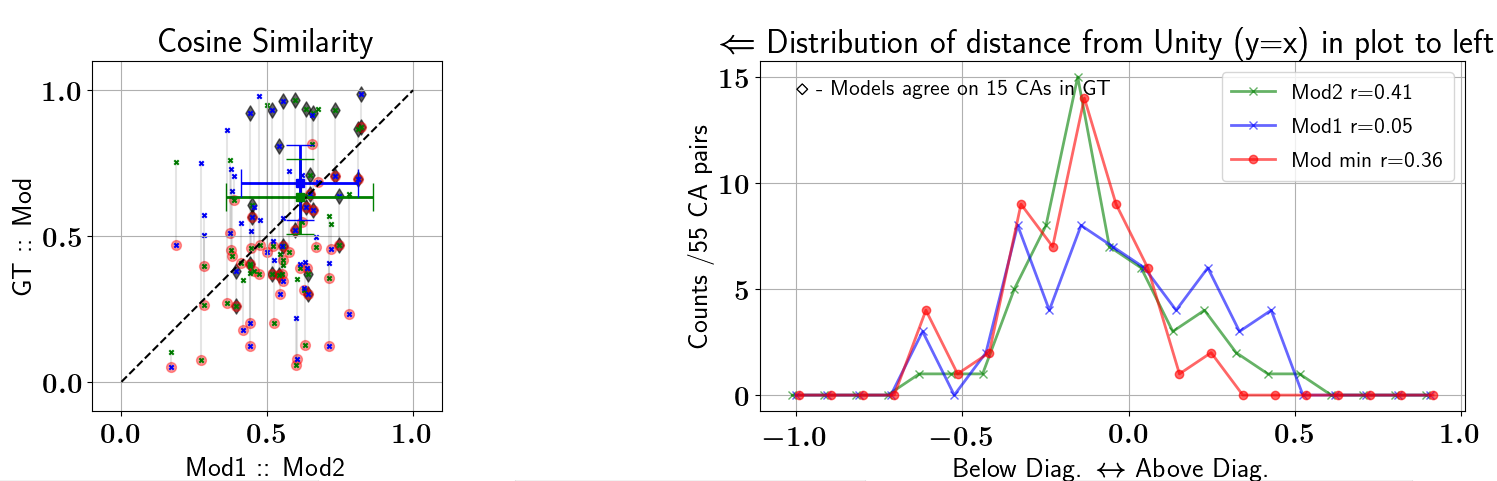}
  \caption{Learning on data fitted to white noise responses}
  \label{fig_singleCAscatter_Wnz}
\end{subfigure}
\vspace*{2mm}
\caption{ \textbf{Individual cross-validated CAs often match GT:} 
Results for synthetic data fitted to natural movie responses in panel \emph{(a)}, and white noise responses in panel \emph{(b)}. Within each panel, the \emph{left plot} compares the cosine similarity ($cs$) between matched pairs of CAs in two learned models (Mod1 and Mod2) on the \emph{x-coordinate} to $cs$ between each model's CA and matching CA in the GT on the \emph{y-coordinate}.
Each point represents similarities between one CA in a model (Mod1 in blue and Mod2 in green) and a matching CA in the GT. The points are connected by a faint grey vertical line if CAs in models are matched to different GT CAs. Red 'o' highlights CA in model pair with smaller $cs$ match to GT, and a black diamond marks points for which the two models are matched to the same CA in the GT (i.e., where green and blue points coincide). Larger blue and green '+' show $\mu$ and $\sigma$ of each model's CA population. Within each panel the \emph{right plot} shows the distribution of distances in the left plot between points and the unity line.  
Pearson correlation coefficients ($r$) for three scatter groups in left plots are shown in legend on the right.}
\label{fig_singleCAscatter}
\end{figure}

Next, we investigate at the level of individual cell assemblies whether agreement between cross-validated models correlates with GT. Extending the analysis begun in Fig~\ref{cosSimPerm}, Fig~\ref{fig_singleCAscatter} quantifies the degree to which the $cs$ value between two CAs matched across a pair of models (see examples in plots on the right of panels \emph{a} and \emph{b} of Fig~\ref{cosSimPerm}) correlates with the $cs$ values between each model's CA and the GT (not shown in Fig~\ref{cosSimPerm}).
Within each panel in Fig~\ref{fig_singleCAscatter}, a point in the scatter plot (left) shows $cs$ for matched CA in model pair \emph{x-axis} vs. $cs$ of each model's individual CA matched with CA in GT \emph{y-axis}. See figure caption for further description. 
For the pair of models fitted to synthetic data resembling natural movie responses \emph{(a)}, we see strong correlation for large $cs$ values,
reflected by the points clustered in the upper right.
Averaging across the entire diagonal \emph{(right plot, a)}, we find the distribution of scatter points tightly peaked close to 0 with correspondingly high $r$ values. Moreover, models trained on synthetic data resembling natural movie responses agree on 39/55 CAs in the ground truth (black diamonds in \emph{left} scatter). As expected, the models trained on synthetic data to white noise responses \emph{(b)} show less agreement with each other and GT, reflected by 
wider distributions, lower $r$ values and only 15/55 CAs agreeing in GT (right). This assessment shows that individual cell assemblies found repeatedly when trained on different partitions of a data set are likely in the GT. This result reassures our final step, to apply the BLV model directly on recording data for which GT is not available.

\subsection{Findings with applying the BLV model directly to retinal recordings}\label{realResults}

\begin{figure}[H]
    \centering
    \begin{subfigure}{.33\textwidth}
        \includegraphics[width=\textwidth]{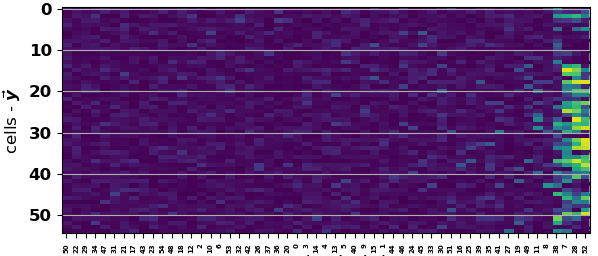}
        \caption{white noise responses}
    \end{subfigure}
    \begin{subfigure}{.32\textwidth}
         \includegraphics[width=\textwidth]{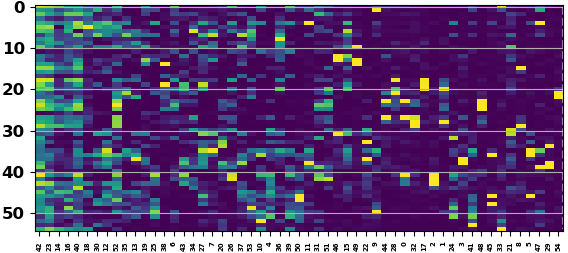}

        \caption{natural movie responses}
    \end{subfigure}
    \begin{subfigure}{.32\textwidth}
          \includegraphics[width=\textwidth]{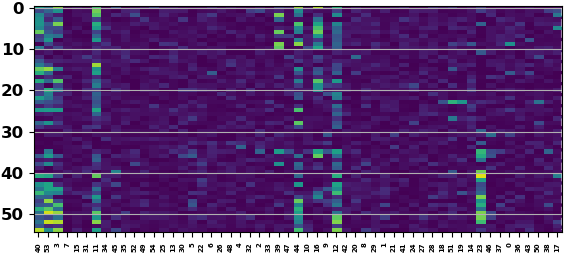}
          \caption{GLM simulated responses}
    \end{subfigure}
    \vspace*{1mm}
    \caption{ \small{ \textbf{Models trained on 55 Off-Brisk Transient responses to different stimuli:} $\mathbf{P}$ matrices for models trained on retinal responses to white noise stimulus \emph{(a)}, retinal responses to natural movie stimulus \emph{(b)} and GLM simulated responses to (same) natural movie stimulus \emph{(c)}. Cell ID on y-axis, CA ID on x-axis. Yellow indicates crisply defined cell membership, i.e., $p(y_i=1|z_a=1) \approx 1$.} }
    \label{example_Pias}
\end{figure}

After training BLV models on the retinal recording data sets described in \ref{realDataSub}, we first examine differences in CA structure between white noise responses vs. natural movie responses. Here we focus on the recordings of 55 offBT RGC responses to both stimulus types. In addition, we also fit a GLM model to the cell responses of the natural movie stimuli  and train a BLV model on these data. 
Fig~\ref{example_Pias} shows typical $\mathbf{P}$ matrices learned from the three data sets in BLV models. Similar results are found for models trained on responses from [offBT,offBS] and [offBT,onBT] cell-type combinations (data not shown). We find that cell assembly structure discovered in natural movie responses is qualitatively different from those in white noise responses and GLM simulated natural movie responses. 
On the experimental natural movie responses, the BLV model finds a large set of CAs with a large fraction of CAs in which the membership structure is crisp and sparse \emph{(b)}. In contrast, on white noise responses \emph{(a)} and GLM fitted natural movie responses \emph{(c)}, the found sets of discriminable CAs are rather small and do not contain crisply defined sparse CAs.  

\begin{figure}[H]
    \centering
    \begin{subfigure}{.315\textwidth}
      \centering
      \includegraphics[width=\textwidth]{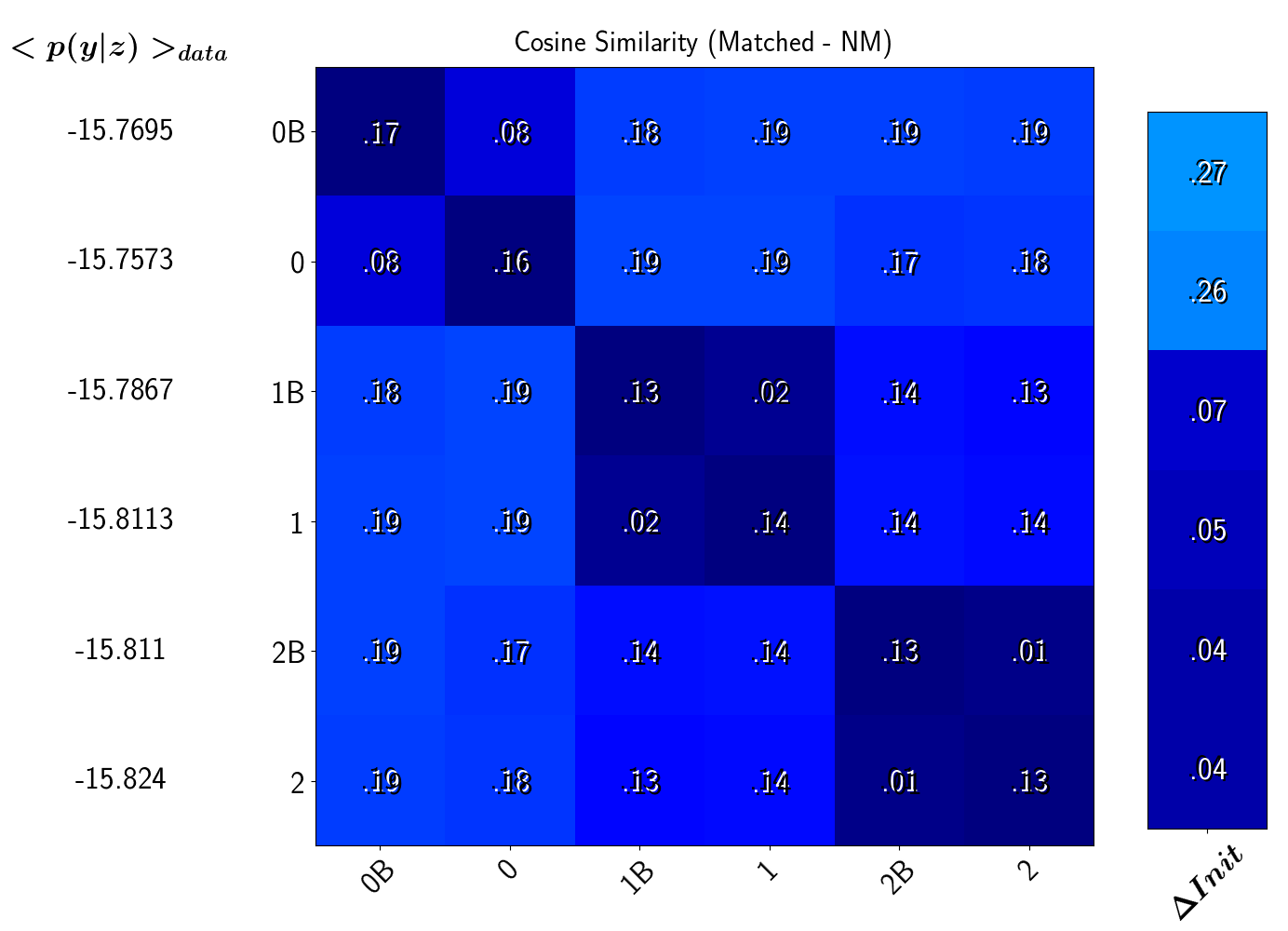}
      \caption{white noise responses}
    \end{subfigure}
    \begin{subfigure}{.315\textwidth}
      \centering
      \includegraphics[width=\textwidth]{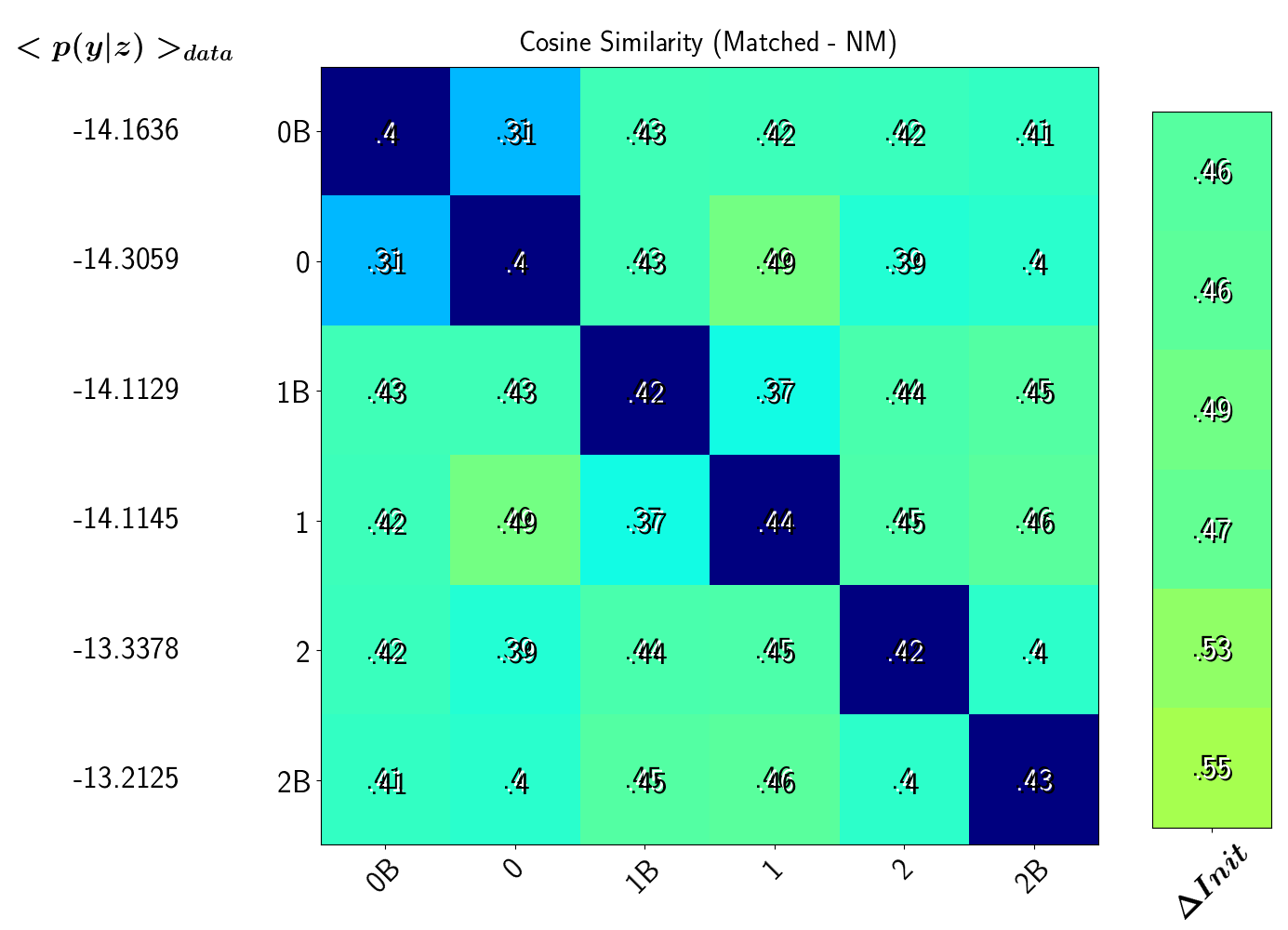}
      \caption{natural movie responses}
    \end{subfigure}
    \begin{subfigure}{.35\textwidth}
      \centering
      \includegraphics[width=\textwidth]{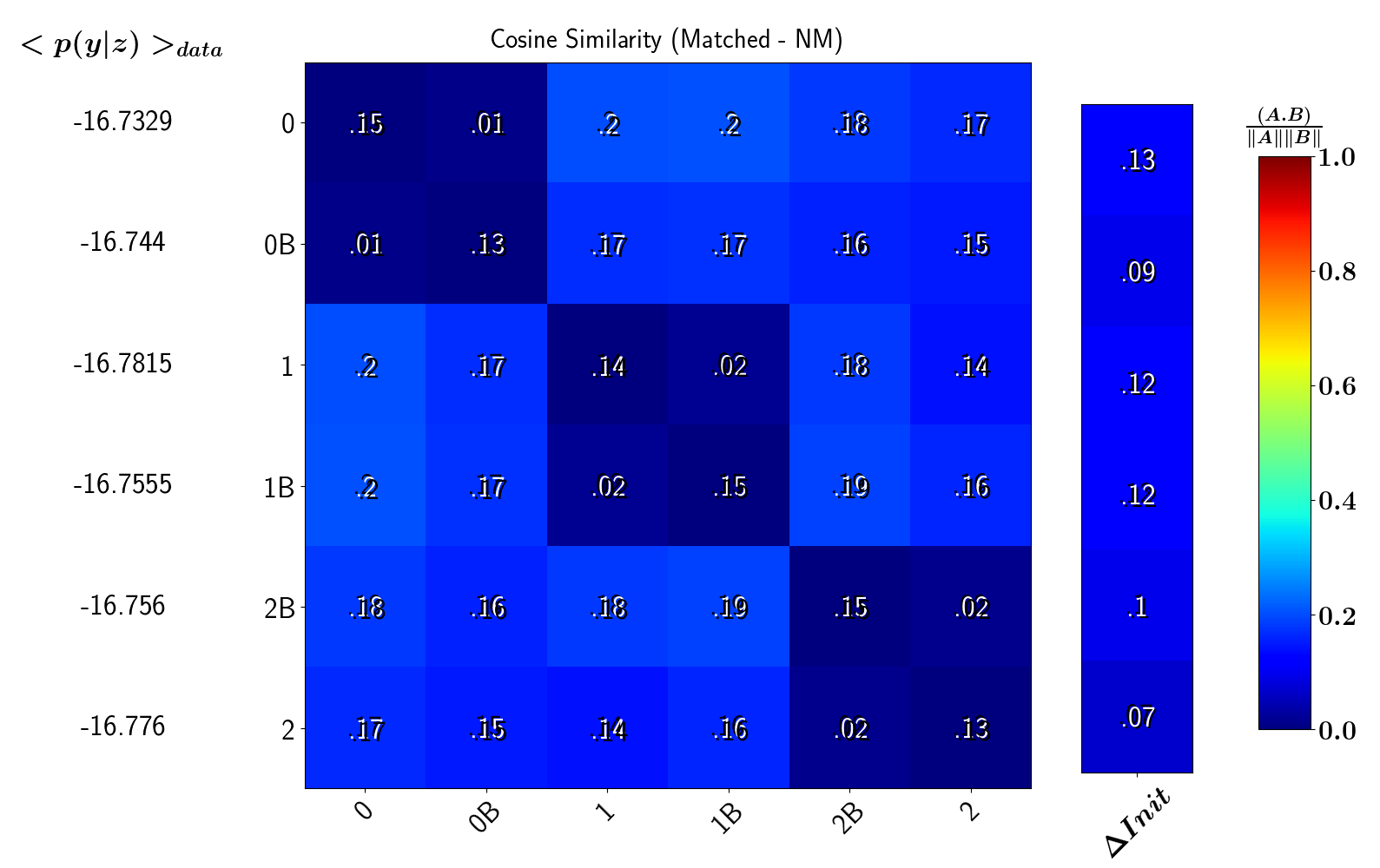}
      \caption{GLM simulated responses}
    \end{subfigure}
    \vspace*{3mm}
    \caption{ \small{ \textbf{Cross-validation of models trained on recorded retinal responses:} Display similar to Fig~\ref{fig_Xval_cosSim}. Similarity of CA membership structure across model pairs for six models trained on white noise retinal responses \emph{(a)}, natural movie responses \emph{(b)}, and GLM simulated responses to same natural movie stimulus \emph{(c)}. Within each panel, matrix off-diagonal elements show average $\Delta cs$ (relative to null without CA matching) between all matched CA pairs within a model pair. Here diagonal value indicates average between a model and all other models, i.e. the average across a row. Numbers on left show average conditional log probability computed on hold out set of half of all spike-words, i.e., cross-validation, for each model. Vector on right shows average change from initialization for all CAs in model, defined as $1-cs$.} }
    \label{csMatrix_models}
    \end{figure}

Following the cross-validation method described in section \ref{synthResults}, we train six models for each data set, displaying the results in Fig~\ref{csMatrix_models}.
Comparison of \emph{(b)} to panels \emph{(a)} \& \emph{(c)} suggests that the models trained on natural movie responses have extracted cell assembly structure that is confirmed by cross-validation while models trained on the other two data types have not.  First, training on natural movie responses changes a much larger fraction of the random initialization of the $\mathbf{P}$ model parameters than the training on the other data sets (compare vectors $\Delta$Init). Second, models independently trained on different partitions of natural movie responses share more structure, indicated by higher values in the matrix off diagonals \emph{(b)}. Third, the average conditional probability of the data in the cross-validation set, $p(\mathbf{y} \vert \mathbf{z})$, is highest for models trained on natural movie responses. These results suggest that meaningful CA structure is discovered in natural movie responses.

\begin{figure}[H]
        \centering
        \includegraphics[width=.93\textwidth]{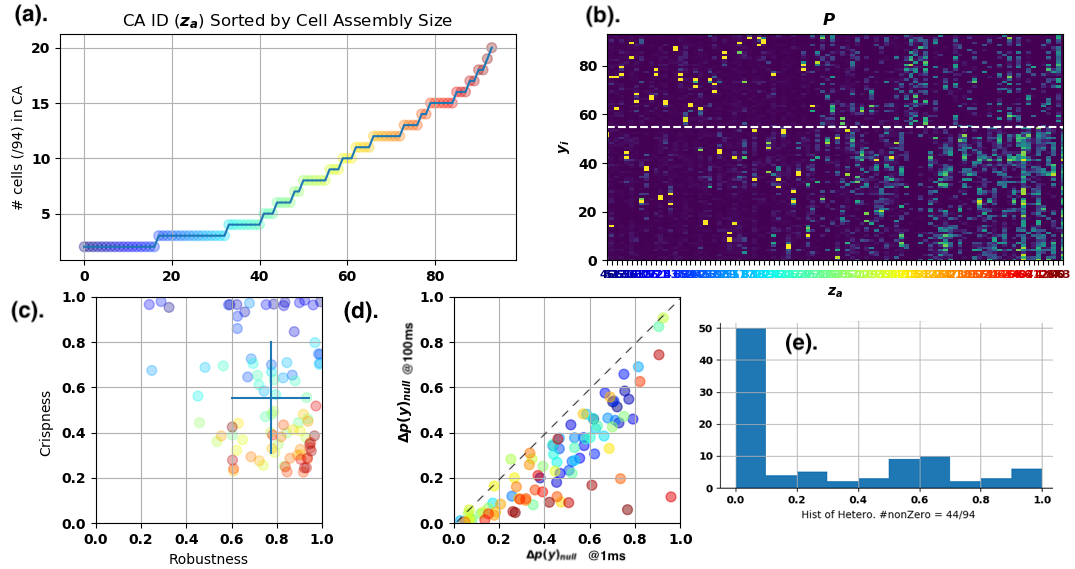}
        \vspace*{1mm}
        \caption{ \small{ \textbf{Statistics of CA metrics in typical model} trained on natural movie responses from 94 [offBT,onBT] RGCs. \emph{(a)}. Sorted CA sizes, color relationship to CA size maintained in CA id in \emph{(b)} and scatter points in \emph{(c)} \& \emph{(d)}. 
        \emph{(b)}. $\mathbf{P}$ matrix. The dashed white line separates offBT cell population (below line) and onBT population (above line). \emph{(c)}. Membership Crispness $C_M$ vs. Cross-validation Robustness $R_X$, each point a CA. Cross shows $\mu$ \& $\sigma$ across all CAs. \emph{(d)}. Difference from null $\Delta P_y$ with 1ms binning vs. with 100ms binning. \emph{(e)}. Heterogeneity $H$ metric histogram.} }
        \label{CA_stats}
    \end{figure}

We now apply the metrics developed in \ref{metrics} to catalog cell assemblies discovered in retinal responses to natural movie and showcase some interesting examples of CAs. The BLV model reveals cell assembly structure that is reliable in terms of cell membership and trial-by-trial activation and that cannot be explained by feed-forward stimulus effects modeled by our null model, the GLM model.
    The statistics of these metrics for $94$ CAs in one typical model trained on [offBT, onBT] responses to a natural movie are shown in Fig~\ref{CA_stats}. The statistics look similar for other models trained on this population, and qualitatively similar for models trained on [offBT] responses and [offBT,offBS] responses to natural movie (data not shown). The CA sizes range from $2$ to $20$ cells. In models trained on [offBT] responses alone, with 55 cells, the maximum size was around a dozen cells, distributions of CA sizes resembling \emph{(a)}, with proportionally more small and crisp CAs. Sorting and coloring CAs by size \emph{(a)} reveals that membership crispness, $C_M$, is correlated with CA size, resulting in the vertical color gradient \emph{(c)} and qualitative difference from left to right in the sorted $\mathbf{P}$ matrix \emph{(b)}. Many CAs are robustly learned across models \emph{(c)} and significantly different from GLM predictions on fine and coarse time-scales \emph{(d)}. Finally, although the majority of CAs do not cross cell-type boundaries, a number of heterogeneous CAs are found as well \emph{(e)}.

\begin{figure}[H]
        \centering
        \begin{subfigure}{.32\textwidth}
        \includegraphics[width=\textwidth]{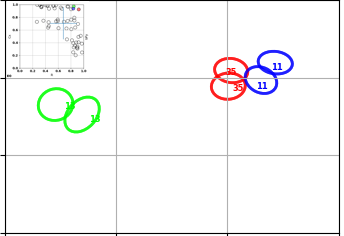}
        \caption{Crisp ($C_M>0.9$)} 
        \end{subfigure}
        \begin{subfigure}{.31\textwidth}
        \includegraphics[width=\textwidth]{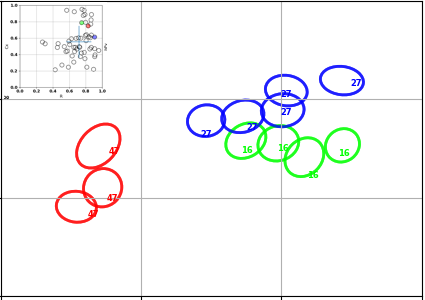}
        \caption{($0.8>C_M>0.6$)}
        \end{subfigure}
        \begin{subfigure}{.31\textwidth}
        \includegraphics[width=\textwidth]{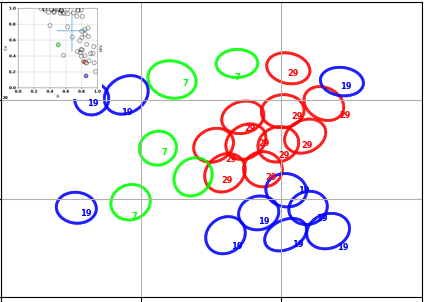}
        \caption{Diffuse ($C_M<0.5$)}
        \end{subfigure}
       
       \vspace*{1mm} \caption{\textbf{ Membership Crispness $C_M$ examples:} Receptive fields of cells from CAs projected to the visual field. We show 9 example CAs from 3 separate models trained on [offBT] natural movie responses. In each panel, colored ovals show RFs of cell members from 3 separate CAs with similar $C_M$ values. Small inset scattters $R_X$ on \emph{x-axis} vs. $C_M$ on \emph{y-axis} for all CAs in each model with shown CAs highlighted in matching color.}
        \label{RvCx_offBT1}
    \end{figure}

    Next we describe CA membership crispness, cross-validation robustness and statistical significance relative to GLM null model predictions within a single cell-type population. In responses from the [offBT] population, we find a variety of CAs from crisp nearest neighbor pairs \emph{(a)} to large diffuse groups \emph{(c)} in Fig~\ref{RvCx_offBT1}. 
    
\begin{figure}[H]
        \centering
        \begin{subfigure}{\textwidth}
        \includegraphics[width=\textwidth]{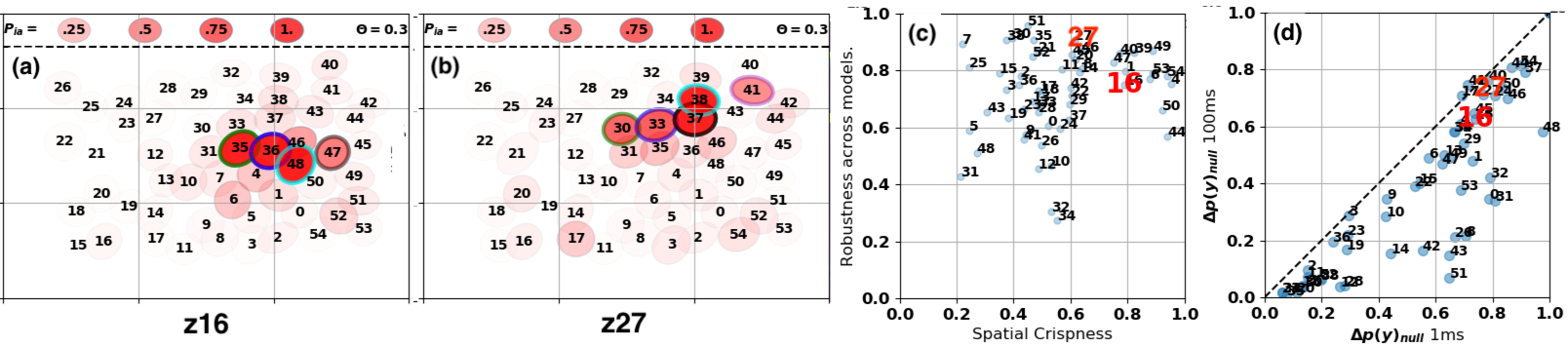}
        \end{subfigure}
        \begin{subfigure}{\textwidth}
        \includegraphics[width=\textwidth]{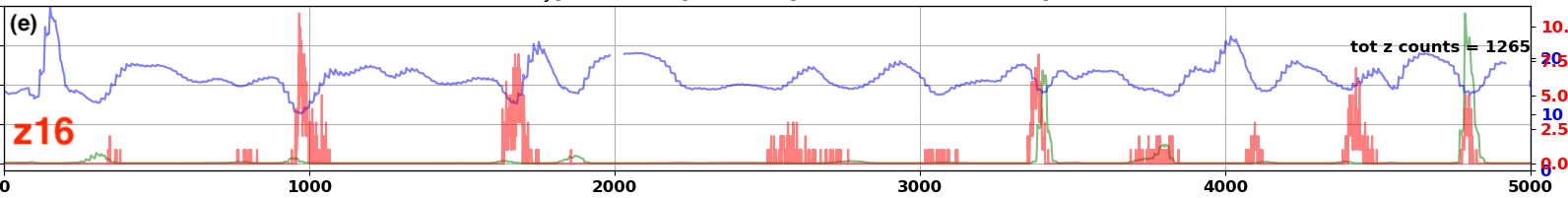}
        \end{subfigure}
        \begin{subfigure}{\textwidth}
        \includegraphics[width=\textwidth]{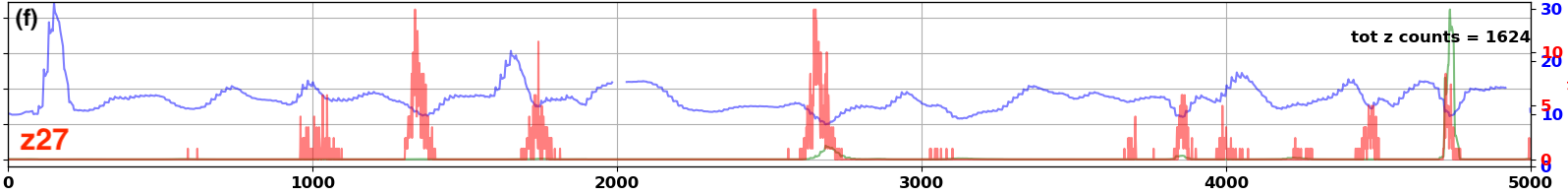}
        \end{subfigure}
       
       \vspace*{1mm} \caption{\textbf{Properties of CAs neighboring in visual space:} Receptive Fields from CAs $z16$ \emph{(a)} and $z27$ \emph{(b)}. Membership crispness ($C_M$) vs cross-validation robustness ($R_X$) for all CAs in model,  two shown highlighted in red \emph{(c)}. Differnce from GLM null predictions $\Delta P_y$ at fine (1ms) vs coarse (100ms) time scales  \emph{(d)}. Temporal response traces of CA inference, PSTH($z_a$) \emph{red}, and $p(\mathbf{\mathbf{y}}_{null}) $ predictions from GLM \emph{green} \emph{(e)} \& \emph{(f)}.}
        \label{z16and27}
    \end{figure}

    Fig~\ref{RvCx_offBT1}\emph{b} shows that CAs $z16$ (green) and $z27$ (blue) each consist of a group of $4 - 5$ neighboring cells. The receptive fields form elongated, horizontally oriented shapes in the visual field.  
    Fig~\ref{z16and27} displays more detailed properties of CAs $z16$ and $z27$. Panels \emph{(a)} and \emph{(b)} show that the group of crisp members in the two CAs are surrounded by weaker members with some cells shared by both CAs.  Both CAs are robust under cross-validation paradigm \emph{(c)}. $\Delta P_y$ computed at 1ms vs 100ms time resolutions \emph{(d)}, respectively, indicate that both CAs are involved in activity significantly different from GLM predictions at coarse and fine time scales. Temporal response traces \emph{(e)} \& \emph{(f)} reveal that even though the CA RFs are proximal in the visual field, share horizontal orientation and some cells, the temporal activations differ significantly between the two CAs \emph{in red}, as well as between the GLM predictions \emph{in green}. 
    
\begin{figure}[H]
        \centering
        \includegraphics[width=.93\textwidth]{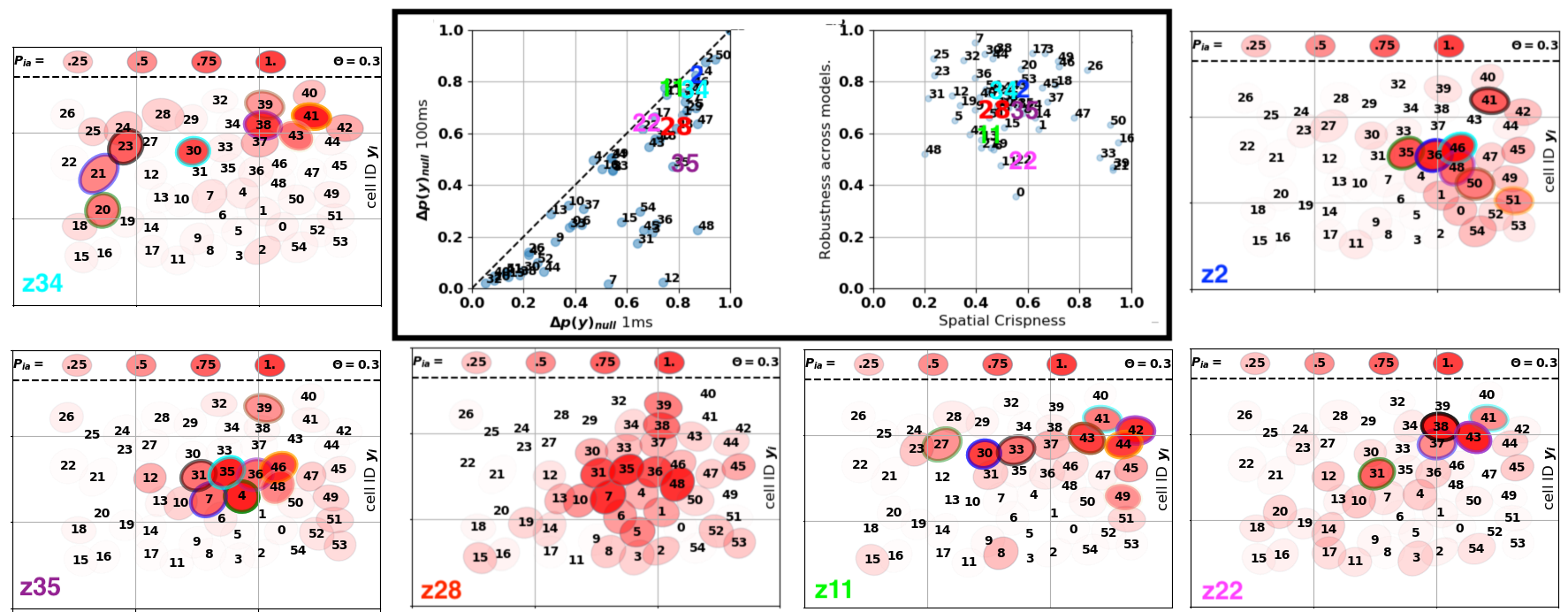}
       
       \vspace*{2mm} \caption{\textbf{Selection of six robust and significant CAs:} 
       Scatter plots in upper center box (bolded) show metric values ($\Delta P_y$, $C_M$, $R_X$) for each CA in model. For the CAs highlighted by color, each of the surrounding panels shows the RFs of the [offBT] member cells. Saturation of red reflects strength of membership in the CA (legend above dashed line). CA id is displayed by colored number in the bottom left of each panel.}
        \label{rand0B_compile}
    \end{figure}

Fig~\ref{rand0B_compile} showcases six of the more than a dozen potentially interesting CAs found by a single model with high cross-validation robustness and significance relative to the GLM-null model at coarse and fine timescales. While the CAs featured have modest crispness values ($0.4<C_x<0.6$), visual inspection reveals reasonably discernible boundaries in terms of cell membership. Moreover, several CAs form elongated shapes in the visual field (specifically $z34$, $z35$, $z11$), possibly encoding extended edges in the stimulus. The resulting neural code is likely a temporal code,  given that the activity in cell assembly members is more synchronous than predicted by the GLM rate-code model. 

    \vspace*{-2mm}
    
\begin{figure}[H]
        \centering
        \begin{subfigure}{.36\textwidth}
          \centering
          \includegraphics[width=\textwidth]{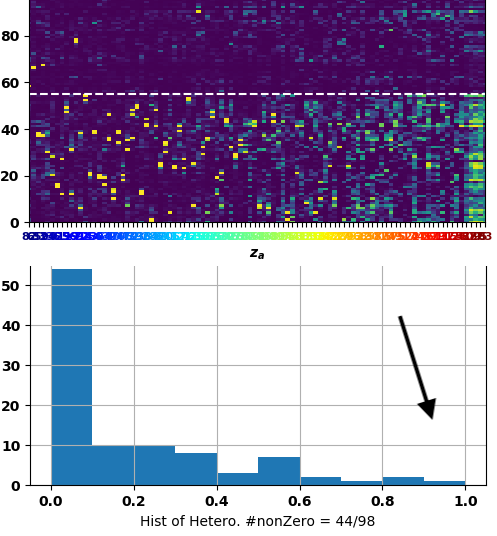}
          \caption{[offBT,offBS]}
        \end{subfigure}
        \hspace{5mm}
        \begin{subfigure}{.36\textwidth}
          \centering
          \includegraphics[width=\textwidth]{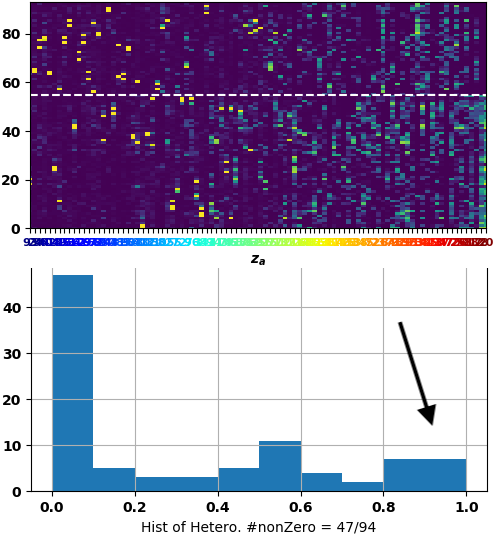}
          \caption{[offBT,onBT]}
        \end{subfigure}
        \vspace*{2mm}
        \caption{ \textbf{Heterogeneity of CAs across cell-types:} BLV models trained on natural movie responses from 55 offBT and 43 offBS RGCs \emph{(a)} and responses from 55 offBT and 39 onBT RGCs \emph{(b)}. In each panel, $\mathbf{P}$ matrices shown on top with columns indicating CAs. OffBT cells below dashed white line. Bottom shows histogram of $H$ metric values for all CAs in model. Black arrow indicates larger number of heterogeneous CAs in [offBT, onBT] population.}
        \label{HeteroHist}
    \end{figure}

    Our method also can equally be applied to identify cell across mixed populations of retinal cell-types responding to natural movie stimulus.
    To investigate whether cell assemblies can involve members of different cell types, we train BLV models on responses from multiple cell types. Fig~\ref{HeteroHist} contrasts two typical results using mixed cell-types, one result for a data set with [offBT, offBS] cells, the other for a data set with [offBT, onBT] cells. For each data set, the 5 other cross-validation models trained resemble those shown. In both data sets, the majority of CAs learned segregate within one or the other cell type. In responses of [offBT, onBT] cells, we find a larger fraction of heterogeneous CAs indicated by the slight shift in the distribution to higher heterogeneity values in panel \emph{(b)} relative to \emph{(a)}. Though subtle, this trend towards more heterogeneous CAs crossing [offBT,onBT] cell-types is consistent across other models learned.

\begin{figure}[H]
        \centering
        \includegraphics[width=.95\textwidth]{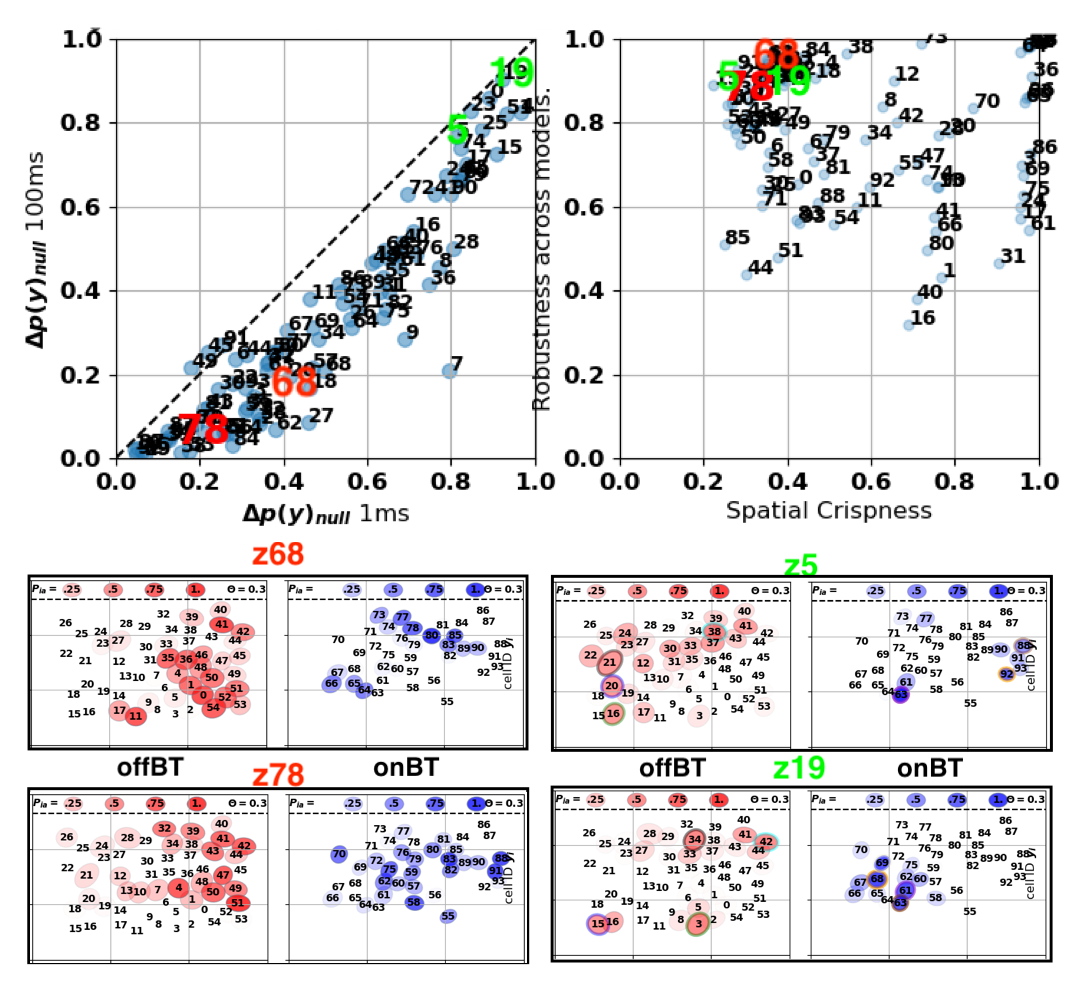}
        \caption{ \textbf{Heterogeneous CAs in [offBT,onBT] population:} Each boxed plot (bottom panels) represents one CA, red ovals showing offBT cell RFs and blue ovals, onBT cells. The upper left scatter plot shows $\Delta P_y$ at 1ms and 100ms time resolutions for all detected CAs. Note in this plot that cell assemblies $z5$ \& $z19$ (green) are among the CAs, most significantly different from GLM null prediction, while $z68$ \& $z78$ (red) are among the least significant CAs. The upper right scatter plot shows $R_x$ vs. $C_M$ for all detected CAs -- the shown heterogeneous CAs (colored numbers) are among the most robust CAs across learned models, but also among the CAs with weakest membership crispness.}
        \label{Hetero_Ex}
    \end{figure}

    We find a number of heterogeneous CAs within the [offBT,onBT] populations. They tend to form extended regions of ON and OFF cells that border one another, perhaps performing some sort of push-pull computation or edge enhancement. Fig~\ref{Hetero_Ex} shows four heterogeneous CAs learned in a single model. All are learned robustly across models and are moderately crisp, \emph{right top}, however, only two of them are significantly different from null model predictions, \emph{left top}. 
    
\begin{figure}[H]
\centering
\begin{subfigure}{\textwidth}
  \centering
  \includegraphics[width=\textwidth]{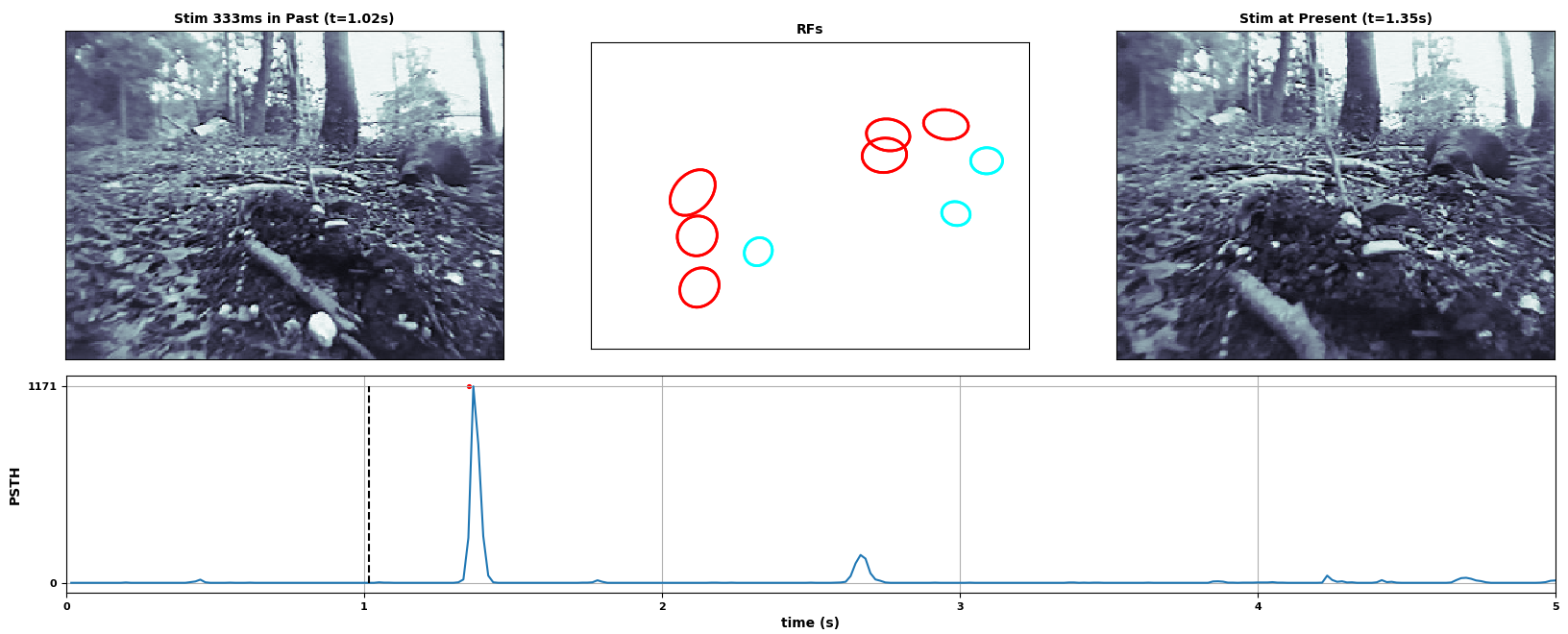}
  \caption{Cell Assembly $z5$ and stimulus:}
  \label{HeteroStim_good1}
\end{subfigure}
%
%
\begin{subfigure}{\textwidth}
  \centering
  \includegraphics[width=\textwidth]{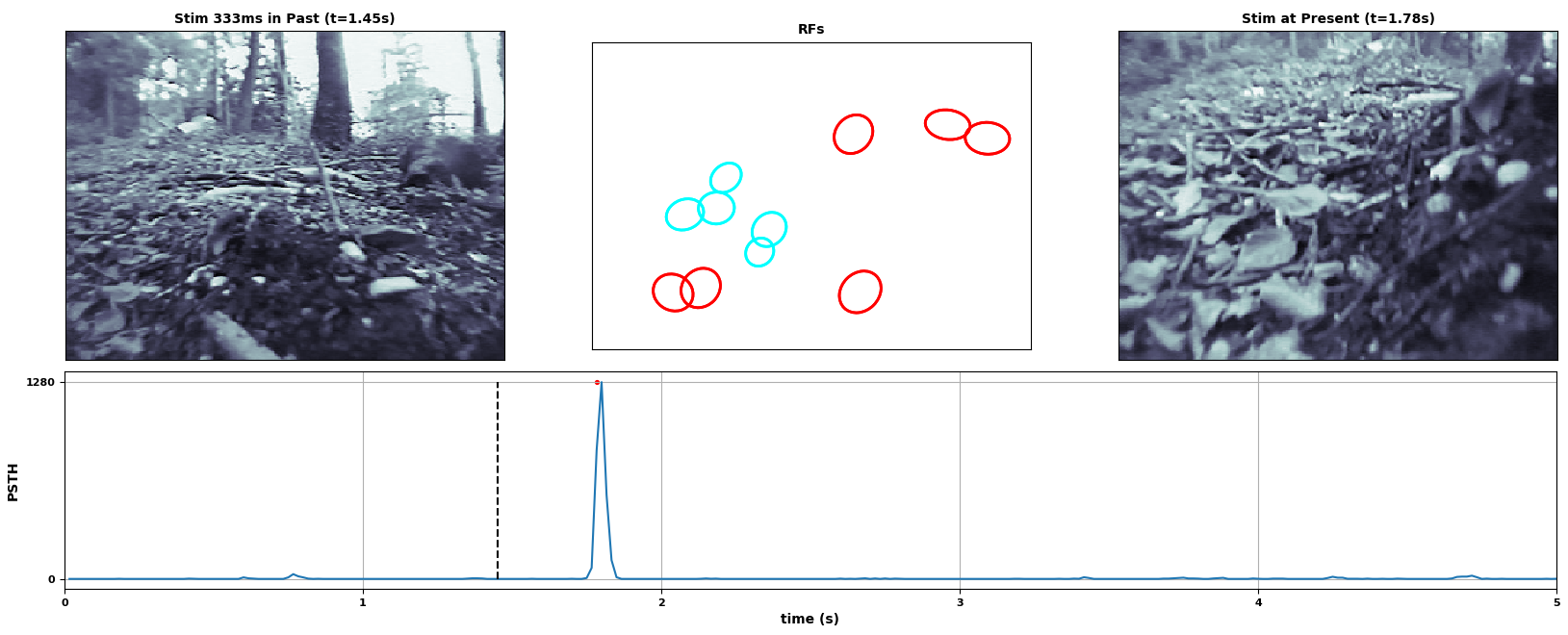}
  \caption{Cell Assembly $z19$ and stimulus} 
  \label{HeteroStim_good2}
\end{subfigure}
\vspace*{2mm}
\caption{ \textbf{Two CAs \emph{significantly different} from GLM null model:} Within each panel, \textit{Center top} shows cell RFs of offBT (red) and onBT (cyan) member cells. 
\textit{Bottom} shows PSTH of CA activation. \textit{Right top} shows stimulus at time of CA activation, blue peak in PSTH. \textit{Left top} shows stimulus $\sim333$ms prior to CA activation, dashed line.}
    
\label{HeteroStim_good}
\end{figure}

     Last, Fig~\ref{HeteroStim_good} presents PSTHs and stimulus frames co-occuring with strong activations of the two heterogeneous CAs that are 
     significantly different from the null prediction, CA  $z5$ in \emph{(a)} and $z19$ in \emph{(b)}. 
     Each CA shows precise and repeatable activation across trials in \emph{bottom} PSTH trace.
     It is unclear how to explain in isolation the spatial configurations of the two types of strongly participating member cells of the two heterogeneous CAs, shown in \emph{top center} plots. These configurations roughly match high contrast structure in the stimulus frames around the time of CA activations but a quantitative analysis of stimulus properties that trigger these CAs is left to future work.

\section{Discussion} \label{Discussion}

    In this work, we introduced a novel probabilistic latent variable model, the BLV model, to detect group structure in binary data. The new model extends the "Noisy-OR model" \cite{heckerman1990} to allow individual variability in observation vectors, adding the $\mathbf{R}$ parameters. Further the Bernoulli prior on latent activation of the Noisy-Or model is extended to a "Homeostatic Egalitarian" prior, yielding in some cases a more even use of the available latent variables. The new model is related to binary soft-clustering, which assigns data points in a high dimensional binary space to multiple cluster centers, that are learned by the algorithm \cite{peters2013}. The model is also related to non-linear sparse coding \cite{olshausen1996}, with the difference, that in our method both observed and latent variables are binary, not real-valued.
    We have investigated several variants of the BLV model, differing in the priors for the latent representations, and vetted them on synthetic data.

    The BLV model was applied to retinal spike-trains recorded from cells responding to white noise and natural movie stimulus. In addition to applying the model directly on the recording data, it was also applied to synthetic data with known causal structure, whose statistics were matched to neural responses to different types of stimuli. 
    Our method was able to discover larger and more crisp cell assemblies in synthetic data matched to responses of natural movies, as compared to synthetic data matched to responses to Gaussian noise. Moreover, each latent cell assembly had lower probability of being active at any one time, relative to cell assemblies learned on synthetic Gaussian noise responses. We validated how consistently the method found cell assemblies in the synthetic data fitted to real spike-trains, finding that structure in the model fitted to natural movie responses was more easily learned. Finding that ground truth CA structure embedded into the data was robustly learned across multiple cross-validated models, we developed some assessment tools which we could then apply to models trained on real data.
    
    When applied directly on the neural recording data, the BLV model revealed cell assembly structure in retinal spike trains not captured by traditional model of retinal encoding, such as GLM models. The CA structure we found was strongly dependent on the type of stimulation. While little CA structure was found in responses to white noise, our method discovered large numbers of robust cell assemblies of various sizes and shapes in retinal responses to natural movies.  The crispest CAs often included a few cells which were nearest neighbors in the receptive field mosaic. The RFs of other cell assemblies formed elongated edges and curves in the mosaic, the least crisp CAs formed diffuse large clusters. 
    
    Using the BLV model to analyze responses across different types of retinal ganglion cells revealed the following results. A large fraction of CAs were entirely homogeneous, exclusively including one cell type. However, we also found CAs that were heterogeneous. For example, a few CAs included both offBT and onBT ganglion cells and aligned intriguingly with structure in the movie stimulus shortly before activation. The temporal response properties of CAs seemed to correlate with their size and complexity, with  PSTHs from smaller pairwise CAs looking similar to single cell PSTHs and larger more complex CAs being activated precisely at one time in the stimulus. Importantly, spike-words observed during many of the CA activations had extremely low probability under an independent GLM model learned on the same data.
    
    The extent of the analysis which could be performed on the retinal data provided was limited however due to several properties of the experimental data -- which were collected before our method was available. Both stimulus types essentially consisted of only 150 image frames and therefore it was infeasible to even performing simple reverse correlation of latent activity with the stimulus. For fully leveraging our method, future experiments should be performed with diverse longer natural movie or naturalistic stimulus without trial repeats. 

    What are the potential questions that can be addressed, using the BLV model in combination with appropriately designed experiments? Inspired by the work of \cite{deny2017} and \cite{koepsell2009} and building on the work discussed in \cite{warner2020}, a question that, for example, could be addressed is how complex natural scenes, correlated in both space and time, are encoded by retina. The experiment to address this question would collect responses to a variety of natural or simple naturalistic movies. Scenes should contain objects that move both laterally and in depth, move relative to one another, and occasionally occlude one another. Specifically, the stimulus should include frames when nearby cell RFs process a common segment and frames when the same set of nearby cells is separated by an image segment boundary. Contrasts and textures should be varied through out the data set to provide a rich and challenging assortment of complex scenes to parse. Including large and abrupt shifts in the visual scene which mimic eye movements would provide insight into how the retina uses or ignores large bursts of activity at stimulus onset or just after a fixation. The experimental data and the results of the CA analysis could then be compared with predictions of the image segmentation model of retina described in \cite{warner2020}.
    
    Other approaches have previously used latent variable models to analyze spiking data, for example, restricted Boltzmann machines (RBMs) \cite{koster2014}.  In contrast to an RBM, the BLV model is a directed graphical model, a causal model of the data where the latent variables represent causes of spike-words. We are not aware of earlier approaches using directed graphical models to analyze neural data. The BLV model is also different from another popular spike analysis method, Unitary Events Analysis (UEA) \cite{grun2010}.  The BLV model learns a real-valued probabilistic representation of cell assemblies and allows an observed spike-word to be represented by a combination of latent variables. This facilitates the detection of noisy repeats of commonly occurring patterns. In contrast, UEA detects only exact repeats of binary patterns and requires many trial repeats to elicit repeat responses. The BLV model does not require stimulus repeats and, in fact, suggests future data collection without stimulus repeats in order to more fully sample the space of natural images and drive the retinal cell population in a wider variety of ways. 
    
    Finally, we wish to reiterate that while we have demonstrated our method on retinal data, the method can find cell groups that fire in synchrony in any neural data. It can also be applied to any other binary data where hidden causes affect subsets of variables. Of course, it should also be kept in mind that the method is agnostic of mechanistic cause. In the case of spike train analysis, the synchrony could be caused by common input, recurrent excitation or any other mechanism. 



\section*{Acknowledgement}
We wish to thank Dr. Greg Field of Duke University for collecting and providing the retinal data, as well as for fruitful discussions and feedback on the analysis. The work was supported by NIH grant 1R01EB026955.

\newpage

\bibliographystyle{abbrv}
\bibliography{Warner_PLOS_CellAssembly.bib}



\end{document}